\documentclass[sn-mathphys,Numbered]{sn-jnl}% 

\usepackage{graphicx}%
\usepackage{multirow}%
\usepackage{amsmath,amssymb,amsfonts}%
\usepackage{amsthm}%
\usepackage{mathrsfs}%
\usepackage[title]{appendix}%
\usepackage{xcolor}%
\usepackage{textcomp}%
\usepackage{manyfoot}%
\usepackage{booktabs}%
\usepackage{algorithm}%
\usepackage{algorithmicx}%
\usepackage{algpseudocode}%h
\usepackage{listings}%
\begin{document}

\title[3D exploration-based search for multiple targets using a UAV]{3D exploration-based search for multiple targets using a UAV}

\author*{\fnm{Bilal} \sur{Yousuf}$^*$}\email{bilal.yousuf@aut.utcluj.ro}

\author{\fnm{Zs\'{o}fia} \sur{Lendek}}\email{zsofia.lendek@aut.utcluj.ro}

\author*{\fnm{Lucian} \sur{Bu\c{s}oniu}$^*$}\email{lucian.busoniu@aut.utcluj.ro}

\affil{\orgdiv{Department of Automation}, \orgname{Technical University of Cluj-Napoca}, \orgaddress{\street{Memorandumului 28}, \city{Cluj-Napoca}, \postcode{400114 }, \state{Cluj}, \country{Romania}}}


\abstract{Consider an unmanned aerial vehicle (UAV) that searches for an unknown number of targets at unknown positions in 3D space. A particle filter uses imperfect measurements about the targets to update an intensity function that represents the expected number of targets. We propose a receding-horizon planner that selects the next UAV position by maximizing a joint, exploration and target-refinement objective.~Confidently localized targets are saved and removed from consideration.~A nonlinear controller with an obstacle-avoidance component is used to reach the desired waypoints. We demonstrate the performance of our approach through a series of simulations, as well as in real-robot experiments with a Parrot Mambo drone that searches for targets from a constant altitude. The proposed planner works better than a lawnmower and a target-refinement-only method.}

\keywords{Multi-target search, unmanned aerial vehicle, probability hypothesis density filter, Parrot Mambo minidrone}

\maketitle

\section{Introduction}

We consider an unmanned aerial vehicle (UAV) exploring a 3D environment to find an unknown number of static targets at unknown locations as quickly as possible, while avoiding obstacles. The UAV is equipped with an imperfect sensor that may miss targets and takes noisy measurements of the detected targets.

Many methods exist to find a known or unknown number of static or dynamic targets from uncertain measurements \cite{Ivic:2022, chen:2022, Aniket:2021, Jun:2020, Pappas:2019, tok:2017, Dam:2015, Good:2014, Haus:2011}. In most of these methods, the path of the sensor is chosen based on the mutual information (MI) between its trajectory and the event of not seeing any targets \cite{chen:2022, Aniket:2021, Jun:2020, Pappas:2019, tok:2017, Dam:2015}. By maximizing this MI, the chances of seeing targets that were previously observed are increased. However, MI-based approaches will focus on those targets that happen to be detected first, failing to discover distant targets. Moreover, most existing techniques concentrate on 2D search.

Therefore, in this paper we propose a novel method for target search that (a) includes an explicit exploration objective in the motion planner, so as to find targets anywhere in the environment and (b) works in 3D space. The only prior work we found that included exploration as an objective is \cite{Fei:2010}, but that approach only works for multiple agents, some of which are designated as exploratory, so it cannot be applied to a single agent that must balance exploration with refining existing targets. Furthermore, we include obstacle avoidance in the UAV control.
% Furthermore, (c) we propose an alternative to MI that maintains performance but is computationally much cheaper, and include obstacle avoidance in the UAV control.

% For example, to find future poses for multiple robots, \cite{Dam:2015,tok:2017} define target-search motion planners based on mutual information (MI). MI measures the amount of information ~A distributed Voronoi-based algorithm is used by \cite{Phi:2020} to drive the robot's motion in coverage and search tasks. \cite{Good:2014} present two hierarchical path planning algorithms (Top2 and TopN) in which the UAV visits high-priority subregions. \cite{Fei:2010} present a cooperative tracking framework using multiple agents, among which several are exploring agents.

The UAV sensor has a limited field of view, modeled using a position-dependent probability of detection at each step. Based on noisy measurements of those targets that are detected, a Sequential Monte Carlo-Probability Hypothesis Density (SMC-PHD) filter \cite{Dou:2005} is run in the framework of random finite sets~\cite{Phi:2020, Mich:2014}. The filter uses  weighted particles to represent an intensity function -- a generalization of the probability density that integrates not to a probability mass but to the expected number of targets \cite{Phi:2020}.

The new UAV path planner picks future positions by maximizing an objective function with two components: exploration and target refinement. The proposed exploration component drives the UAV toward unseen regions of space, and is implemented using an exploration bonus initialized to $1$ over the entire environment and then decreased in observed regions. Target refinement aims to better locate targets about which measurements were previously received, and is computed in one of two ways: MI, like in \cite{chen:2022, Aniket:2021, Jun:2020, Pappas:2019, tok:2017, Dam:2015}, as well as a computationally cheaper alternative in which the probabilities of detection at estimated target locations are summed up.~As a result of maximizing the objective function, the planner returns a sequence of positions, the first of which is implemented using a backstepping controller that includes an obstacle-avoidance component for known obstacles. The procedure is repeated in receding horizon.

Estimated target locations are computed as the centers of K-means clusters of particles \cite{Tuo:2014, Bear:2006}. Narrow enough clusters that contain a large enough intensity mass are declared as found targets.~To prevent reidentifying found targets, corresponding particles and future measurements likely to be originating from them are removed, using a new method, different from those in \cite{Aniket:2021, Phi:2020}.

The proposed planner is extensively validated in a series of simulations and experiments.~We first study the influence of the horizon and find that horizon 1 suffices in practice.~We also illustrate that the algorithm finds more targets over time than a lawnmower and an MI-only method.~We compare MI to our computationally cheaper center-probability technique and find no statistically significant difference in target detection.~The real-life experiment involves a Parrot Mambo mini-drone exploring at a constant altitude, in which the method works better than a lawnmower, confirming the simulation results.

This paper is a heavily extended and revised version of our early conference version \cite{Bil:2022}, with the following extra contributions: (i) the algorithm is upgraded from 2D to 3D, (ii) the planner is defined in receding horizon whereas it was myopic (single-step) before; 
% (iii)  we include the new center-probability refinement component; 
(iii) an obstacle avoidance strategy is added to the control of the UAV, and, importantly, (iv) the framework is validated with real-life experiments using a Parrot Mambo mini-drone. The conference version \cite{Bil:2022} was also extended -- along a different line from this paper -- to multiagent search in \cite{Bil:2023}, which is restricted to simulations in the 2D case, but does include center-probability refinement.
% In contrast to the work of other researchers discussed at the start of the section, our method includes exploration and works in 3D.

Next, Section \ref{sec:problem} formulates the problem, followed by background on SMC-PHD in Section \ref{sec:filter}.  Section \ref{sec:exploration} describes the proposed method. Section \ref{sec:simulation} presents simulation results, followed by the hardware setup and experimental results in Section \ref{sec:experiment}. Section \ref{sec:conclusion} concludes the paper.
\section{Problem Formulation}\label{sec:problem}

\begin{figure}[!htb]
  \centering
  \includegraphics[scale=0.22]{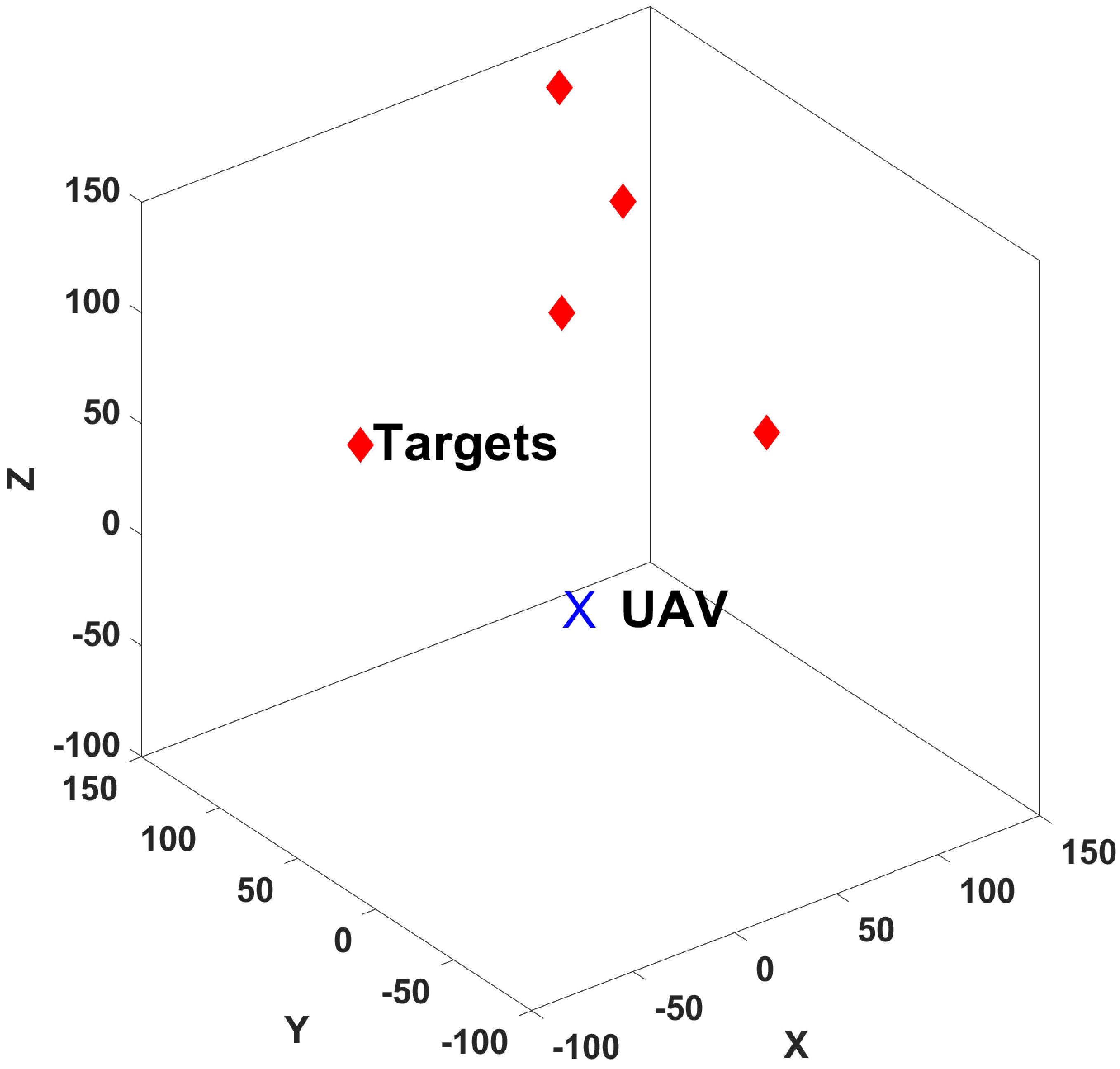}
  \includegraphics[scale=0.2]{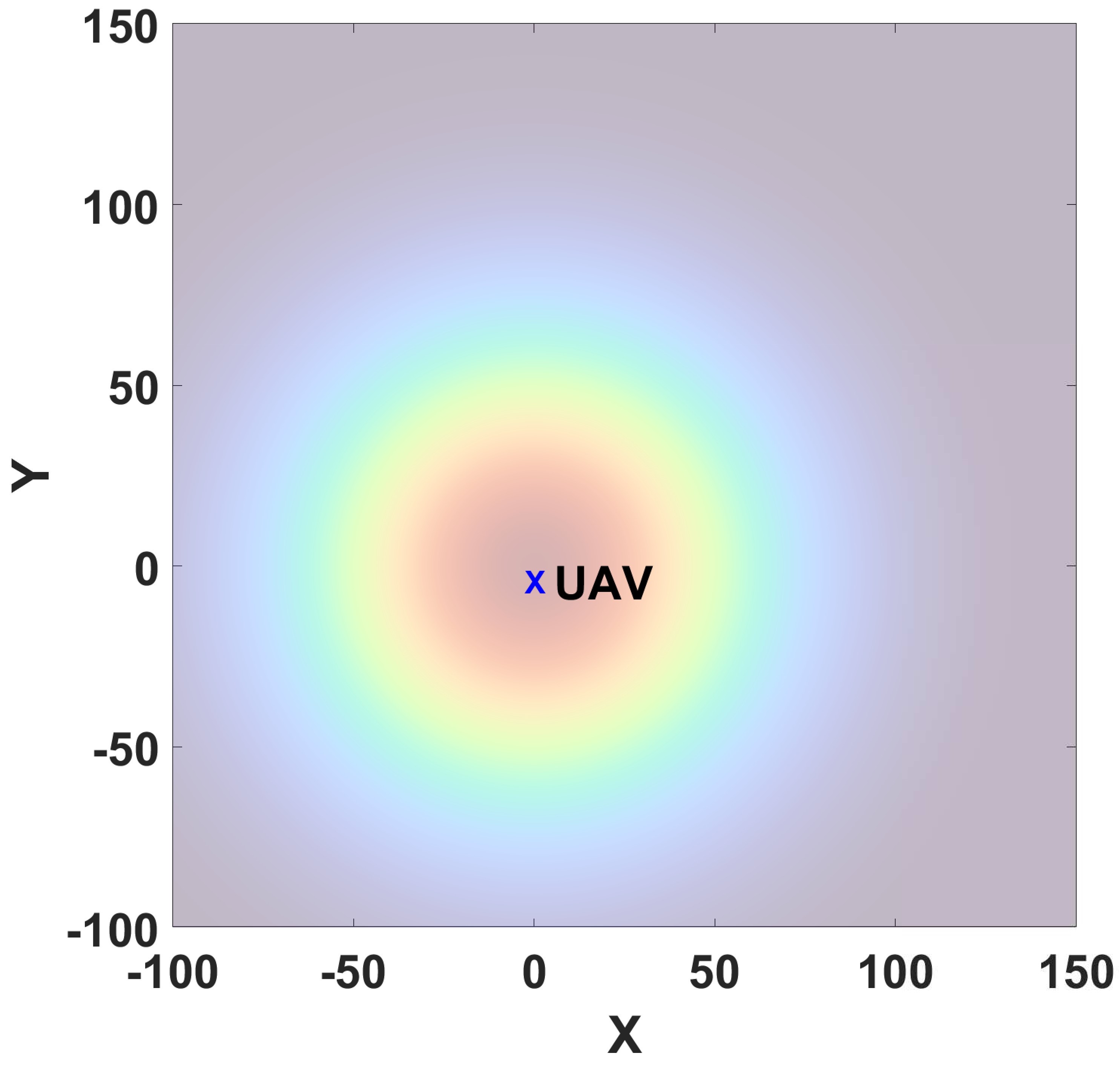}
  \caption{Left: 3D space with 5 targets and a UAV. Right: UAV with a spherical field of view symmetrical in all three axes. The dark orange to blue colors show the probability of observation (higher to lower) at the current position of the UAV. This is a 2D slice of the 3D probability of observation.}
  \label{Fig:scene}
%  \vspace{-1em}
\end{figure}

The problem we consider is a UAV that explores a bounded 3D space (environment) $E$ in search of an unknown number of static targets, as illustrated in Fig.~\ref{Fig:scene} (left). The main objective is to find the position of all the targets in as few steps as possible.

The position $q\thinspace \in \thinspace E$ of the UAV is assumed to be sufficiently accurate to disregard its uncertainty. The set $X_{k}$ contains $N_{k}$ stationary targets at positions $x_{ik} \thinspace \in  \thinspace E$, where $k$ denotes the discrete time step. Both the cardinality $N_{k}$ and the locations of the targets are initially unknown. Note that although the real set of targets is static, a time-dependent notation is still useful to estimate their positions, since new targets are seen over time. % in the field of view (FOV) of the UAV.

The sensor model describes how measurements occur based on the actual targets. Sensor uncertainty manifests in two ways. Firstly, at each step, the sensor does not see all the targets, but may miss some, depending on a probabilistic field of view. Secondly, the measurements are affected by noise for those targets that are seen. Note that the general formalism  also supports clutter, i.e.\ false target detection, but we will not be using this feature here.

In general, the probability with which the UAV at position $q$ detects a target at position $x$ is denoted by $\pi(x, q)$. Here, we model an omnidirectional ranging sensor for which the probability is defined as:
\begin{equation}\label{eq:pi}
\pi(x,q)= Ge^{-\left\|\zeta\right\|/2}
\end{equation}
where scalar $G \leq 1$ and:
$$\zeta = \left(\frac{\mathcal{X}_{x}-\mathcal{X}_{q}}{\mathbb{F}_{\mathcal{X}}},\frac{\mathcal{Y}_{x}-\mathcal{Y}_{ q}}{\mathbb{F}_{\mathcal{Y}}},\frac{\mathcal{Z}_{x}-\mathcal{Z}_{q}}{\mathbb{F}_{\mathcal{Z}}}\right)
$$
is a normalized distance between the target and the sensor (UAV). In this expression, ${\footnotesize(\mathcal{X}_{x},\mathcal{Y}_{x},\mathcal{Z}_{x})}$ is the 3D position of the target,
${(\mathcal{X}_{q},\mathcal{Y}_{q},\mathcal{Z}_{q})}$ are the 3D coordinates in the UAV position $q$ defined above, and $(\mathbb{F}_{\mathcal{X}},\mathbb{F}_{\mathcal{Y}},\mathbb{F}_{\mathcal{Z}})$ are normalization constants that may be interpreted as the size of the (probabilistic) field of view. For example, when these constants are all equal, $\pi$ is radially symmetric around the UAV position, as illustrated in Fig.~\ref{Fig:scene} (right).

The binary event $b_{ik}$ of detecting a target $x_{ik}$ then naturally follows a Bernoulli distribution given by the probability of detection at $k$: $b_{ik} \sim \ss(\pi(x_{ik},q_{k}))$. Given these Bernoulli variables and the actual target positions $(\mathcal{X}_{x_{ik}},\mathcal{Y}_{x_{ik}},\mathcal{Z}_{x_{ik}})$ in the space $E$, the set of measurements $Z_{k}$ is:
\begin{equation}\label{eq:sensor_model}
Z_{k}=\bigcup_{i\thinspace\in\left\{1,\hdots,N_{k}\right\}\thinspace \mathrm{s.t}. \thinspace b_{ik}=1}\left[h_{ik}(x_{ik})+\varrho_{ik}\right]
\end{equation}
where $h_{ik}(x_{ik})$ is defined as:
\begin{equation*}
\begin{aligned}
h_{ik}(x_{ik})=&\left[d_{ik}, \theta_{ik}, \varpi_{ik}\right]^{T}\\
d_{ik}=& {\footnotesize \sqrt{(\mathcal{X}_{x_{ik}}-\mathcal{X}_{q_{k}})^2+(\mathcal{Y}_{x_{ik}}-\mathcal{Y}_{q_{k}})^{2}+(\mathcal{Z}_{x_{ik}}-\mathcal{Z}_{q_{k}})^{2}}},\\
\theta_{ik}=&{\footnotesize \arctan\frac{\mathcal{Y}_{x_{ik}}-\mathcal{Y}_{q_{k}}}{\mathcal{X}_{x_{ik}}-\mathcal{X}_{q_{k}}}},\thinspace\thinspace
\varpi_{ik}={\footnotesize \arcsin\frac{\mathcal{Z}_{x_{ik}}-\mathcal{Z}_{q_{k}}}{d_{ik}}}
\end{aligned}
\end{equation*}
So, for each target that is detected, the measurement consists of a range $d_{ik}$, bearing angle $\theta_{ik}$, and elevation angle $\varpi_{ik}$ with respect to the UAV. This measurement is affected by Gaussian noise
$\varrho_{ik} \sim\mathcal{N}(.,\mathbf{0},R)$, with mean $\mathbf{0} = [0,0,0]^\top$ and diagonal covariance $R = \mathrm{diag}(\sigma^{2}, \sigma^{2}, \sigma^{2})$.

Based on the measurement model, the target measurement density is:
\begin{equation}\label{eq:gaussian}
g(z_{k}|x)=\mathcal{N}(z_{k},h_{k}(x),R)
\end{equation}
where $g(z_{k}|x)$ is a Gaussian density function with covariance $R$ and centered on $h(x)$, the application of $h$ defined above to some arbitrary $x$ rather than the particular target location $x_{i}$ from \eqref{eq:sensor_model}. This density will be used to estimate the target locations.
% -------------------------------------------------------------------------
% ----- SECTION BREAK -------------------------------
% ------------------------------

\section{Background on PHD Filtering}\label{sec:filter}
The Probability Hypothesis Density (PHD) $D:E\rightarrow[0,\infty)$, or intensity function, is similar to a probability density function, with the key difference that its integral $\int_{S} D(x) dx$ over some subset $S \subseteq E$ is not the probability mass of $S$, but the expected number of targets in $S$. % In particular, if $S=E$, the integral is not $1$, but the (expected) total number of targets.
The PHD filter performs Bayesian updates of the intensity function based on the target measurements and is summarized as:
\begin{equation}\label{eq:summerized}
\begin{aligned}
&D_{k|k-1}=\Phi_{k|k-1}(D_{k-1|k-1})\\
&D_{k|k}=\Psi_{k}(D_{k|k-1}, Z_k)
\end{aligned}
\end{equation}
Here, $D_{k|k-1}$ is the prior intensity function predicted based on intensity function $D_{k-1|k-1}$ at time step $k-1$, and $D_{k|k}$ denotes the posterior generated after processing the measurements. The multi-target prior $D_{k|k-1}$ at step $k$ is defined by:
\begin{equation}\label{eq:PHD_predict}
\begin{aligned}
&{\footnotesize D_{k|k-1}(x_{k})=\Phi_{k|k-1}(D_{k-1|k-1})(x_{k})} = \\
&{\footnotesize\Upsilon_{k}(x_{k})+\int_{E} p_{s}(\xi)\delta_{\xi}(x_{k}) D_{k-1|k-1}(\xi) d\xi}
\end{aligned}
\end{equation}
where $p_{s}(\xi)$ is the probability that an old target at position $\xi$ still exists, and the transition density of a target at $\xi$ is defined as the Dirac delta $\delta_{\xi}(x)$ centered on $\xi$, \cite{Dou:2005}, since in our case targets are stationary. Finally, $\Upsilon(x)$ denotes the intensity function of new targets appearing in the FOV.

Now, to compute the posterior intensity function $D_{k|k}$ at step $k$ using the measurements $Z_k$, we apply the PHD posterior operator $\Psi_{k}(D_{k|k-1}, Z_k)(x_{k})$ to the prior intensity function $D_{k|k-1}$:
\begin{equation}\label{eq:PHD_update}
\begin{aligned}
&{\footnotesize D_{k|k}(x_{k})=\Psi_{k}(D_{k|k-1}, Z_k)(x_{k})}=\\
&{\footnotesize\left[1-\pi(x_{k}, q_k)+\sum_{z\thinspace\in\thinspace Z_{k}}\frac{\psi_{kz}(x_{k}) }{\left<\psi_{kz},D_{k|k-1}\right>(x_{k})}\right]}\cdot D_{k|k-1}(x_{k})
\end{aligned}
\end{equation}
where $\psi_{kz}(x_{k})=\pi(x_{k},q_{k})g(z_{k}|x_{k})$ denotes the overall probability density of detecting a target at $x_{k}$ with $g$ defined in \eqref{eq:gaussian}, and  $\left<\psi_{kz},D_{k|k-1}\right>=\int_{E}\psi_{kz}(x_{k})D_{k|k-1}(x_{k})dx_{k}$. In practice, we employ the SMC-PHD filter \cite{Dou:2005}, which uses at each step $k$ a set of weighted particles $(x_{k}^{i}, \omega_{k|k}^{i})$ to represent $D_{k|k}$, with the property that $\int_{S} D_{k|k} (x) dx \approx \sum_{x_{k}^{i} \in S} \omega_{k|k}^{i}$ for any $S \subseteq E$. For more details about PHD filtering and its particle-based implementation, see Appendix \ref{sec:appendix}.  

\begin{figure}[t]
  \centering
  \includegraphics[trim={8cm 0cm 10cm 0},scale=0.2]{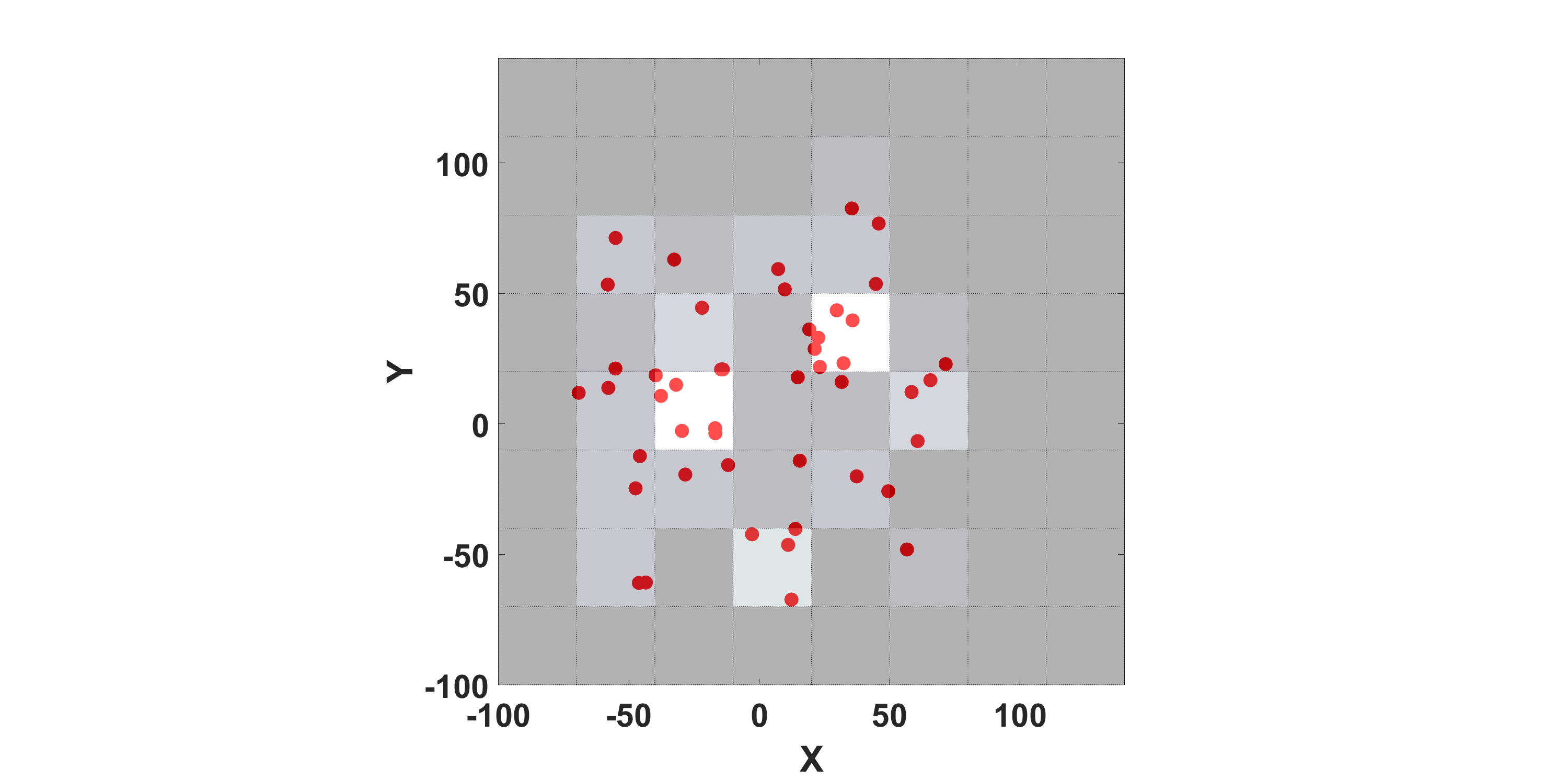}
  \caption{Particle representation and illustration of the corresponding intensity measure in the simpler case of 2D space. The red circles show the particles, and the shades of the grid squares define the expected number of targets present in each square, equal to the sum of the weights of the particles in that square.}
  \label{Fig.particles_filtering}
\end{figure}

An example of an intensity function is given in Fig.~\ref{Fig.particles_filtering}, where to illustrate, the sets $S$ are the grid squares (they could in general have any shape) and the shade of gray is the expected number of targets in each $S$, equal to the sum of the weights of the particles in $S$.

The PHD filter will be used by the planner in the following section to estimate target locations from measurements.

% -------------------------------------------------------------------------
% ----- SECTION BREAK -------------------------------
% ------------------------------
\section{Exploration-Based Search}\label{sec:exploration}

This section presents the main contribution of the paper: the novel target search algorithm, organized in three components. First, the UAV path planner is described in Section \ref{subsec:planner}. Second, in Section \ref{subsec:mark}, we present the method to mark well-defined targets as found and disregard measurements likely to come from these targets in the future. Finally, the UAV controller, including obstacle avoidance, is given in Section \ref{subsec:control}. 

\subsection{Planner}\label{subsec:planner}

Consider first the problem of designing a $3$D path to follow so as to find the targets.~A classical solution to this problem would be a 3D variant of a lawnmower trajectory, which fills the space in a uniform manner.~We evaluate the lawnmower as a baseline in our experiments, but a solution that finds the targets more quickly is desired.~We propose next a receding-horizon path planner that generates such a solution, which, in addition to exploring the space, focuses on refining potential targets that were already (but poorly) measured. %~In contrast to the earlier version of \cite{Bil:2022}, this planner (i) works in 3D, compared to 2D for the earlier approach, (ii) finds a longer-horizon solution, compared to single-step, and (iii) introduces a computationally cheaper variant of the refinement component.

To formalize the planner, define first the integer horizon $\tau > 0$, and a sequence of next positions of the robot $\mathbf{q}_{k} = (q_{k+1}, q_{k+2}, \dotsc, q_{k+\tau})$.~In this sequence, each next position is defined relative to the previous one:
\begin{equation}\label{eq:posegeneration}
q_{k+j+1} = q_{k+j} + \delta q_{j}, \text{ for } j = 0, \dotsc, \tau-1
\end{equation}
where the set of possible position changes $\delta q$ is discrete and should be sufficiently rich to find the targets, see Figure \ref{fig:candidates} for an example.~The planner then solves the following optimization problem:
\begin{equation}\label{eq:optimization}
\mathbf{q}_{k}^{*}\in\mathrm{argmax}_{\mathbf{q}_{k}} \left\{ \alpha\cdot\mathbb{E}(\mathbf{q}_{k}) + \mathbb{T}(\mathbf{q}_{k}) \right\}
\end{equation}

The \emph{exploration component} $\mathbb{E}(\mathbf{q}_{k})$ of \eqref{eq:optimization} is novel, and drives the robot to look at unseen regions of the environment. Define first an exploration bonus function $\iota$, which is initialized to $1$ for the entire environment, and decreases at each step $k$ and each position $x$ by an amount related to $\pi(x, q_{k})$. The meaning is that each position $x$ has been explored to an amount related to the probability of detection at that position. To implement the exploration bonus, we represent $\iota$ on a 3D grid of points $x_{ijl}$, initialized with:
$$
\iota_0(x_{ijl}) = 1, \forall i, j, l
$$
and updated with:
\begin{equation}\label{eq:explbonus}
\iota_{k}(x_{ijl}) = \iota_{k-1}(x_{ijl}) \cdot (1-\pi(x_{ijl},q_{k})), \forall i, j, l, \forall k \geq 1
\end{equation}
Then, the exploration component is defined as:
\begin{equation}\label{eq:expl}
\mathbb{E}(\mathbf{q}_k) = \sum_{j=1}^{\tau} \sum_{i=1}^{c_k} \iota_{k}(q_{k})
\end{equation}
where $\iota_k$ at positions $q_{k}$ that are not on the grid is computed by trilinear interpolation.

The \emph{target refinement} component $\mathbb{T}(\mathbf{q}_{k})$ in \eqref{eq:optimization} focuses on resolving targets about which measurements were already received, by driving the robot to areas where the intensity function is larger.~The refinement component can be computed in two ways.
The first option is the MI between the targets and the empty measurement set along the horizon, which we use and compute as in \cite{Dam:2015}.
%(the procedure is rather intricate, so we will not repeat it here).
Since this MI is maximized, the event of receiving empty measurements is expected to become low-probability. Note that since probabilities of detection depend on the position sequence $\mathbf{q}_{k}$, the MI also depends on these positions.

The second option is developed as a computationally more efficient alternative to MI \cite{Bil:2023}. Specifically, at each step $k$, we extract potential targets as clusters of particles generated with K-means \cite{Dou:2005}. The target refinement component then consists of the sum of the observation probabilities of all cluster centers, accumulated along the horizon:
\begin{equation}\label{eq:co}
\mathbb{T}(\mathbf{q}_{k}) = \sum_{j=1}^{\tau} \sum_{i=1}^{c_k} \pi(\hat{x}_{i,k},q_{k+j})
\end{equation}
where the probability of detection $\pi$ was defined in \eqref{eq:pi}, $c_k$ denotes the number of clusters at $k$, and $\hat{x}_{i,k}$ is the center of the $i$th cluster $C_{i,k}$. The center is denoted by $\hat x$ because it has the meaning of an estimated target position. This option will be called ``center probabilities'', for short. The intuition is that the probability of observing estimated target locations is maximized.

Note that in classical target search \citep{chen:2022, Aniket:2021, Jun:2020, Pappas:2019, Dam:2015}, only MI target refinement is used, which means that the robot only focuses on targets that it already saw, without exploring for new targets. Conversely, when no targets have been seen yet (or when all seen targets have been marked as found, see below for details), planner \eqref{eq:optimization} will compute the robot's trajectory solely based on the exploration component, so a space-filling lawnmower-like trajectory is obtained, see Fig. \ref{Fig.scene7} in the experiments for an example. In most cases, both objectives are important, and the proposed planner strikes a balance between them, controlled by the tuning parameter $\alpha$.

% To bring the UAV to the desired next state $q^*_{k+1}$, a controller must be designed, which is done in the following section.

% -------------------------------------------------------------------------
% ----- SECTION BREAK -------------------------------
% ------------------------------
\subsection{Marking and Removal of Found Targets}\label{subsec:mark}

Even when certain targets are well-determined (i.e.\ they have clear peaks in the intensity function), the target refinement component will still focus on them, which is not beneficial since the time would be better spent refining poorly seen targets or looking for new targets. To achieve this, the algorithm removes such well-determined targets, as explained next. After each measurement is processed, we extract potential targets as clusters of particles with K-means \cite{Dou:2005}. Then, each cluster that is narrow enough, and associated to a large enough concentration of mass in the intensity function, is taken to correspond to a well-determined, or \emph{found}, target. Two conditions are checked: that the cluster radius $r_{i,k}$ is below a threshold $\mathcal{T}_{r}$, and that the sum of the weights $\omega_{j}$ of the particles in the cluster is above a mass threshold $\mathcal{T}_{m}$. The center $\hat{x}_{i,k}$ of such a cluster is added to a set $\hat {X}$ of found targets, and the particles belonging to that cluster are deleted. % we also remove the corresponding particles, and we disregard future measurements likely to come from those targets. <-- duplication!

To prevent the formation of another intensity peak at the locations of old, already found targets, measurements associated to these targets are also removed from the future measurement occurs in sets $Z_k$. Of course, the algorithm has no way of knowing which target generated a certain measurement. Instead, among any measurements $z_{k}$ that are closer than a threshold $\mathcal{T}_{z}$ to some previously found target $\hat x \thinspace\in\thinspace \hat{X}$, i.e., $\left\|h^{-1}(z_{k})-\hat{x}\right\| \leq \mathcal{T}_{z}$, one closest measurement to $\hat x$ is removed from $Z_k$ (note the transformation of $z$ to Cartesian coordinates). Only a single measurement is removed, rather than all measurements closer than $\mathcal{T}_z$, since we know from the measurement model that a target generates at most one measurement.

% -------------------------------------------------------------------------
% ----- SECTION BREAK -------------------------------
% ------------------------------
\subsection{UAV Control Design with Obstacle Avoidance}\label{subsec:control}

The UAV model and the proposed backstepping control law follow \cite{Sun:2014}. The nonlinear model is:
\begin{equation}\label{eq:UAV_model}
\begin{aligned}
\dot{q}=&y(\vartheta)+w\\
\dot{\vartheta}=&f(\vartheta)+\mathbb{g}(\vartheta)u
\end{aligned}
\end{equation}
The state signal consists of two components: $(q,\vartheta) \thinspace \in \thinspace \Re^{7}$. The first component is the already-defined UAV position $q=[\mathcal{X}_{q},\mathcal{Y}_{q},\mathcal{Z}_{q}]$, which is from now on required to be in the north, east, down (NED) frame. The second component, $\vartheta=[V_{a},\beta,\gamma,\phi]^{T} \in \Re^{4}$, contains the airspeed and the heading, pitch, and roll angles, respectively. The control $u=[u_{Va},u_{\beta},u_{\gamma},u_{\phi}]^{T}\thinspace \in \thinspace \Re^{4}$ includes the airspeed rate and angular rates for the three angles, in the same order as in $\vartheta$. The vectors $y$ and $f$ and the matrix $\mathbb{g}$ from \eqref{eq:UAV_model} are:
\begin{equation*}
\begin{aligned}
y(\vartheta)=&\begin{pmatrix}
V_{a}\cos\beta \cos\gamma+w_{n}\\
V_{a}\sin\beta \sin\gamma+w_{e}\\
-V_{a}\sin\gamma+w_{d}
\end{pmatrix},\thinspace\thinspace\thinspace\thinspace f(\vartheta)=\begin{pmatrix}
-\frac{D}{m}-g \sin\gamma\\
-\frac{g}{V_{a}}\cos\gamma\\
0\\
\sin\phi
\end{pmatrix}
\end{aligned}
\end{equation*}
\begin{equation*}
\mathbb{g}(\vartheta)=\begin{pmatrix}
\frac{1}{m} &0  &0 \\
0 &\frac{g}{V_{a}}\cos\phi  &0 \\
0 & 0 &\frac{L}{mV_{a}\cos\gamma}\\
0& 0& 0
\end{pmatrix}
\end{equation*}
where $m$ is the UAV mass, $g$ the gravitational constant, $D$ the drag force, and $L$ the aerodynamic lift. Moreover, $w=(w_{n},w_{e}, w_{d})^{T}$ is the wind (disturbance) vector.

We use the invertibility of functions $\mathbb{g}(\vartheta)$ and $\Im_{y}=\frac{\partial y}{\partial \vartheta}$ to define the tracking control:
\begin{equation}\label{eq:Control_signal}
\begin{aligned}
u=&(\Im_{y}(\vartheta)\mathbb{g}(\vartheta))^{-1}(-\Im_{y}(\vartheta)f(\vartheta)-(\mathcal{K}_{g_{1}}+\mathcal{K}_{g_{2}}+\mathcal{K}_{g_{3}})(\dot{q})\\
&-(1+\mathcal{K}_{g_{1}}\mathcal{K}_{g_{2}}\mathcal{K}_{g_{3}})(q-q^{d}))
\end{aligned}
\end{equation}
where $(\mathcal{K}_{g_{1}},\mathcal{K}_{g_{2}}, \mathcal{K}_{g_{3}})$ are positive gains.

\begin{figure}[hb]
  \centering
  \includegraphics[scale=0.35]{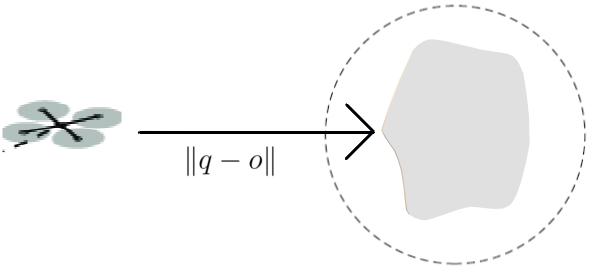}
  \caption{Obstacle avoidance. When the distance between the UAV and the gray obstacle is below a threshold, symbolized by the dashed circle (a sphere in 3D), the avoidance term is applied.}
  \label{Fig:obs_det}
\end{figure}
A simple and computationally efficient method is used to avoid known obstacles. % Each obstacle is enclosed in a ball, of the same size for each obstacle.
A collision danger indicator between the UAV and the obstacle closest to it, see also Fig.~\ref{Fig:obs_det}, is computed with:
\begin{equation}\label{eq:obs_detect}
\mathbb{O}=I\left( \left \| q - o \right \|<d_{l}\right)
\end{equation}
Binary variable $\mathbb{O}$ is equal to 1 when the distance between the position of the UAV $q$ and that of the closest obstacle $o$ is below a collision avoidance distance limit $d_{l}$. The controller equation (\ref{eq:Control_signal}) is modified with a collision avoidance term in the heading control input $u_{\beta}$, which gets activated whenever $\mathbb{O}$ becomes $1$:
\begin{equation}\label{eq:obs_avoid}
\tilde{u}_{\beta}=u_{\beta}-k_{obs}\mathbb{O}
\end{equation}
where $k_{obs}$ is a positive gain.

The control law above is coupled with the planner by using it in-between discrete time steps $k$ and $k+1$ to reach the reference $q^*_{k+1}$ imposed by the planner. An overall piecewise constant reference signal is obtained.

\medskip
Algorithm \ref{algo:horizon} summarizes the entire target search procedure, integrating the components from Sections \ref{subsec:planner}--\ref{subsec:control}.

\begin{algorithm}[!htb]
\caption{Target search at step $k$}\label{algo:horizon}
\begin{algorithmic}[1]
\State use K-means to find clusters $C_{i,k}, i = 1, \dotsc, c_k$
\For{each cluster $i = 1, \dotsc, c_k$}
\If {$r_{i} \leq \mathcal{T}_{r}$ and $\sum_{j \thinspace \in C_k}\omega_{j}\geq \mathcal{T}_{m}$} target found:
\State delete all particles $j \in C_{i,k}$
\State $\hat{X} = \hat{X} \cup \hat{x}_{i,k}$
\EndIf
\EndFor
\State get measurements $Z_k$ from sensor
\For {$\hat{x}\thinspace \in \thinspace \hat{X}$}
\State $Z_\mathrm{aux} = \{z_k \in Z_k \ \vert\  \Vert h^{-1}(z_{k})-\hat{x} \Vert \leq \mathcal{T}_{z}\}$
\If {$Z_\mathrm{aux}$ is nonempty}
\State remove measurement: $Z_{k}=Z_{k} \setminus \mathrm{argmin}_{z_k \in Z_\mathrm{aux}} \Vert h^{-1}(z_{k})-\hat{x}\Vert$
\EndIf
\EndFor
\State run filter from Section \ref{sec:filter} with measurements $Z_{k}$
\State update exploration component $\iota_k$ using \eqref{eq:explbonus}
\For{each $\mathbf{q}_{k}$ generated with \eqref{eq:posegeneration}}
\State compute exploration $\mathbb{E}(\mathbf{q}_{k})$ \eqref{eq:expl} and target refinement $\mathbb{T}(\mathbf{q}_{k})$
\EndFor
\State find best sequence $\mathbf{q}_{k}^*$ with \eqref{eq:optimization}
\State go to the first position $q^*_{k+1}$ of $\mathbf{q}_{k}^{*}$ using the controller \eqref{eq:obs_avoid}
\end{algorithmic}
\end{algorithm}

% -------------------------------------------------------------------------
% ----- SECTION BREAK -------------------------------
% ------------------------------
\section{Simulation Results}\label{sec:simulation}
To validate the efficiency of the proposed target search algorithm, we ran $7$ simulated experiments in 3D target space, referred to as $E1$ through $E7$. In $E1$, we investigate the performance of the receding-horizon planner as a function of the horizon $\tau$. In $E2$ and $E3$ we compare the new algorithm, a standard lawnmower, and a planner that only uses MI without exploration, the latter being a representative baseline of the target search literature \cite{Ivic:2022, chen:2022, Jun:2020, Pappas:2019, Dam:2015, Good:2014}.~$E2$ and $E3$, respectively, concern uniformly distributed and clustered targets.~In $E4$, we evaluate the influence of the maximum cluster width $\mathcal{T}_{r}$ on the errors between the real and estimated targets.~In $E5$, we compare MI with center-probabilities target refinement.~Note that all other experiments employ MI, to align with the existing target search literature.~In $E6$, an experiment is run without any targets to demonstrate how the drone fills the space with an exploratory trajectory similar to a lawnmower.~Finally, $E7$ illustrates the efficacy of our obstacle avoidance scheme.

\begin{figure}[!htb]
  \centering
  \includegraphics[scale=0.4]{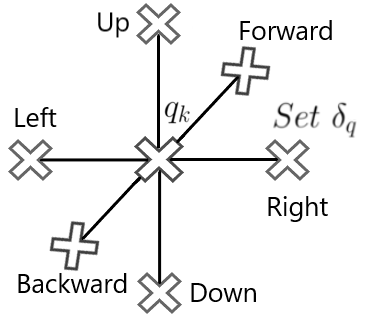}
  \caption{Example set of position changes $\delta q$, relative to the current state which is in the center.}
  \label{fig:candidates}
\end{figure}

In all these experiments, the 3D environment is $E=[-20,260]\thinspace\thinspace\times\thinspace[-20,260]\thinspace\thinspace\times\thinspace[-20,260]$ m$^3$. The distance from the current position at $k$ to the future position at $k+1$ is set to $12$ meters at each step, and the candidates $\delta q$ are six different position choices at this distance, as shown in Fig.~\ref{fig:candidates}. When a 3D lawnmower is used, the altitude difference between each 2D ``lawnmower layer'' is constant and set to 48 meters, and the $X-Y$ lawnmower spacing in each such layer is also set to 48 meters.

The parameters of the probability of detection $\pi(x,q)$ from Section \ref{sec:problem} are: $G=0.98$, $\mathbb{F}_{\mathcal{X}}=\mathbb{F}_{\mathcal{Y}}=\mathbb{F}_{\mathcal{Z}}=25$. We set the maximum number of particles as 5000 to reduce the computation time, and the intensity function for spontaneous birth $\Upsilon_{k}$ is constant and equal to $130$. The threshold values in Algorithm 1 are set as $\mathcal{T}_{r}=1.1$m, $\mathcal{T}_{m}=2.2$, and $\mathcal{T}_{z}=5$m.

The parameters required for the control law \eqref{eq:Control_signal} are selected experimentally to be $\mathcal{K}_{g1}=\mathcal{K}_{g2}=\mathcal{K}_{g3}=9$, and the parameters for the UAV model~\eqref{eq:UAV_model} are adopted from \cite{Sun:2014}: $L=0.7$, $ g=9.8$, $m=10$ and $D=0.9$.~The initial position of the UAV is set to $q_{0}=[0,0,0]$, in with the initial speed, heading, pitch angle, and roll angle listed as $[V_{a},\beta,\gamma,\phi]=[15, \pi/4,0,0]$.~The wind vector is chosen as $[3,0,0]$m/s.~The time required for the UAV using this control to reach the desired future state $q_{k+1}^{*}$ from its current state $q_{k}$ is around $2.4$s.

\begin{figure}[t]
    \centering
    \includegraphics[scale=0.17]{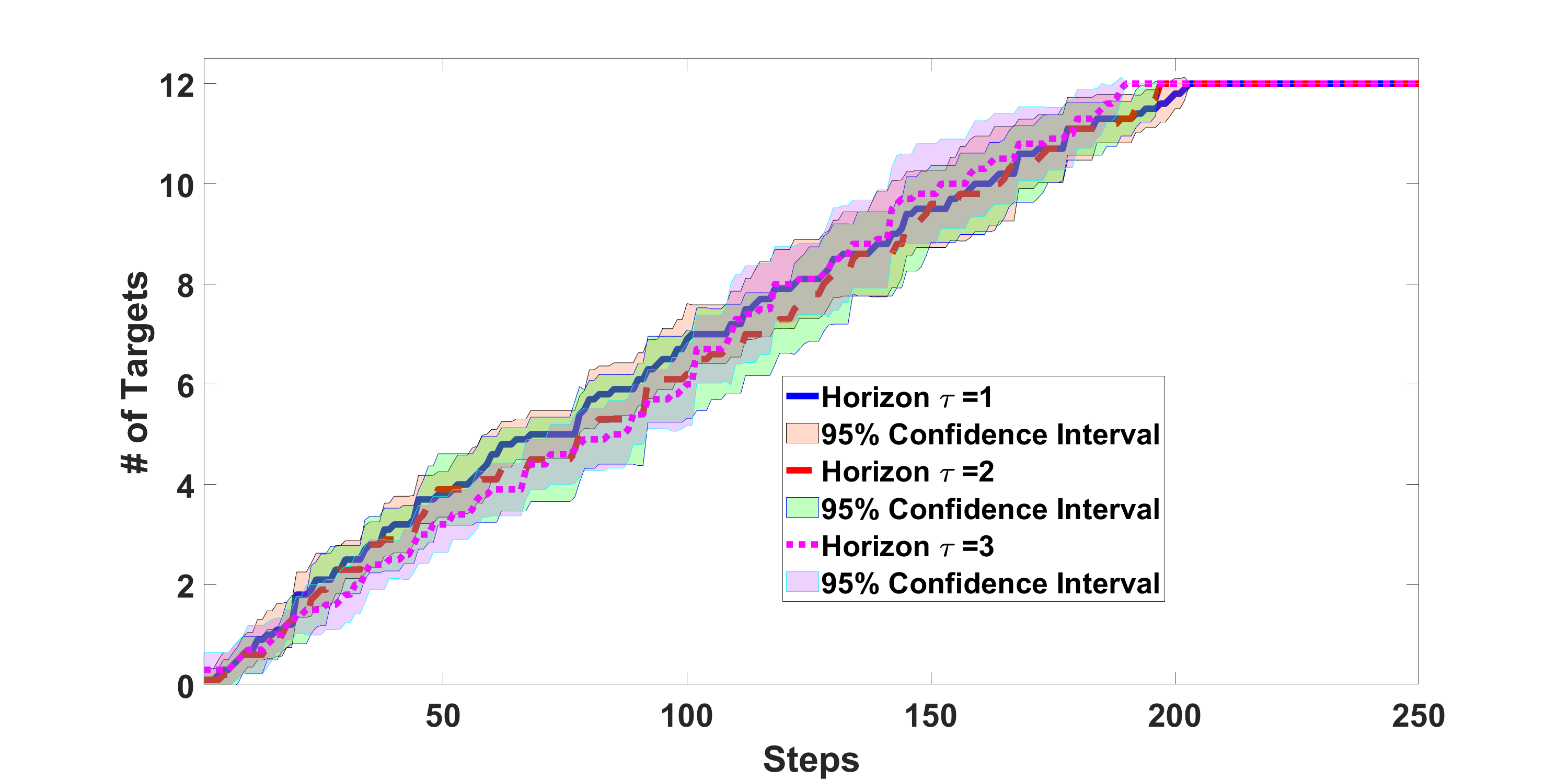}
    \caption{Detected average number of targets placed uniformly at random locations in 10 maps with $95$\% confidence intervals on the mean, using horizons $\tau=[1,2,3]$.}
    \label{Fig.planner}
\end{figure}
\begin{figure}[t]
    \centering
    \includegraphics[scale=0.055]{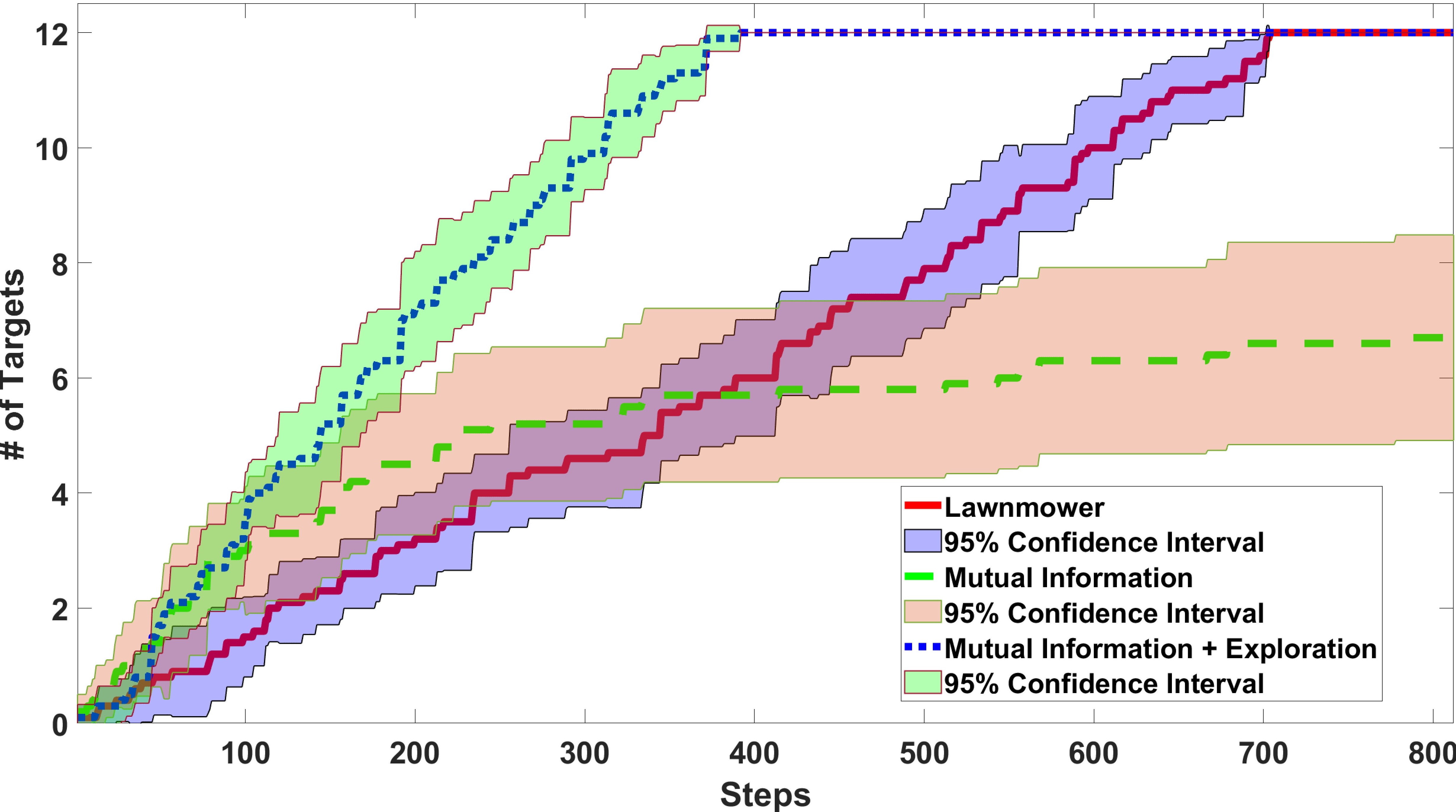}
    \includegraphics[scale=0.07]{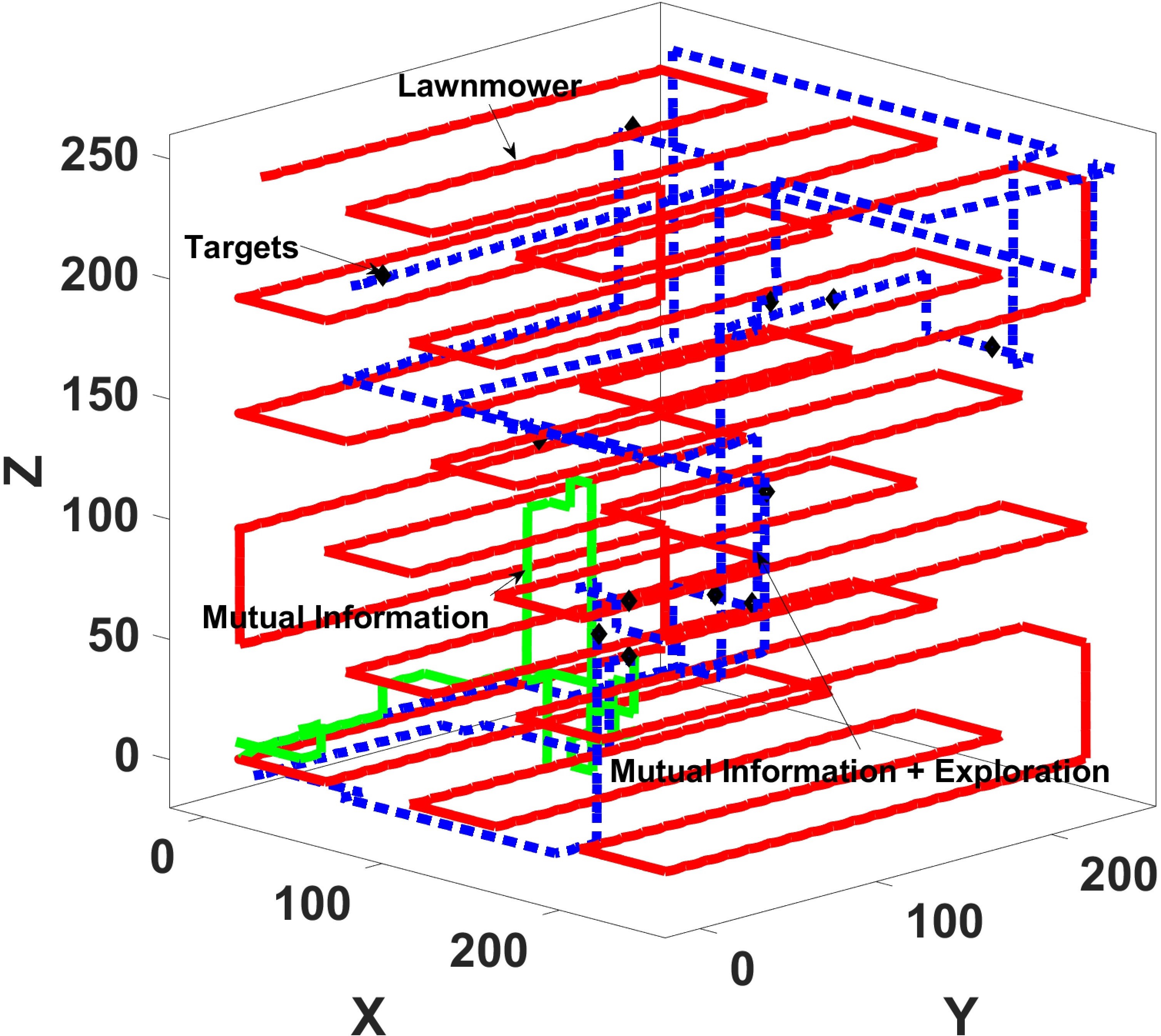}
    \caption{Left: Detected average number of targets on 10 maps with our algorithm, the lawnmower, and the mutual-information-only method. Right:Example trajectories with the three methods.}
    \label{Fig.scene2}
\end{figure}

In experiments $E1-E3$ and $E5$, we report the mean number of targets detected along with $95$\% confidence intervals on the mean out of $10$ independent runs. The trajectory length is chosen differently for each experiment so that all algorithms have a chance to detect all the targets.

\medskip \noindent \textit{E1: Influence of the planner horizon.}~We consider $12$ targets uniformly distributed at random locations. The trajectory length is chosen as $250$ steps. Fig.~\ref{Fig.planner} shows the number of target detections over time, for a varying horizon $\tau=1, 2, 3$.~It can be seen that horizon $1$ is statistically indistinguishable from $2$ and $3$. Therefore, we choose horizon $\tau={1}$ for the remaining simulations as it is computationally more efficient.

\medskip \noindent \textit{E2: Planner performance for uniformly-distributed targets.} We consider $12$ targets uniformly randomly distributed throughout $E$.
\begin{figure}[t]
  \centering
  \includegraphics[scale=0.16]{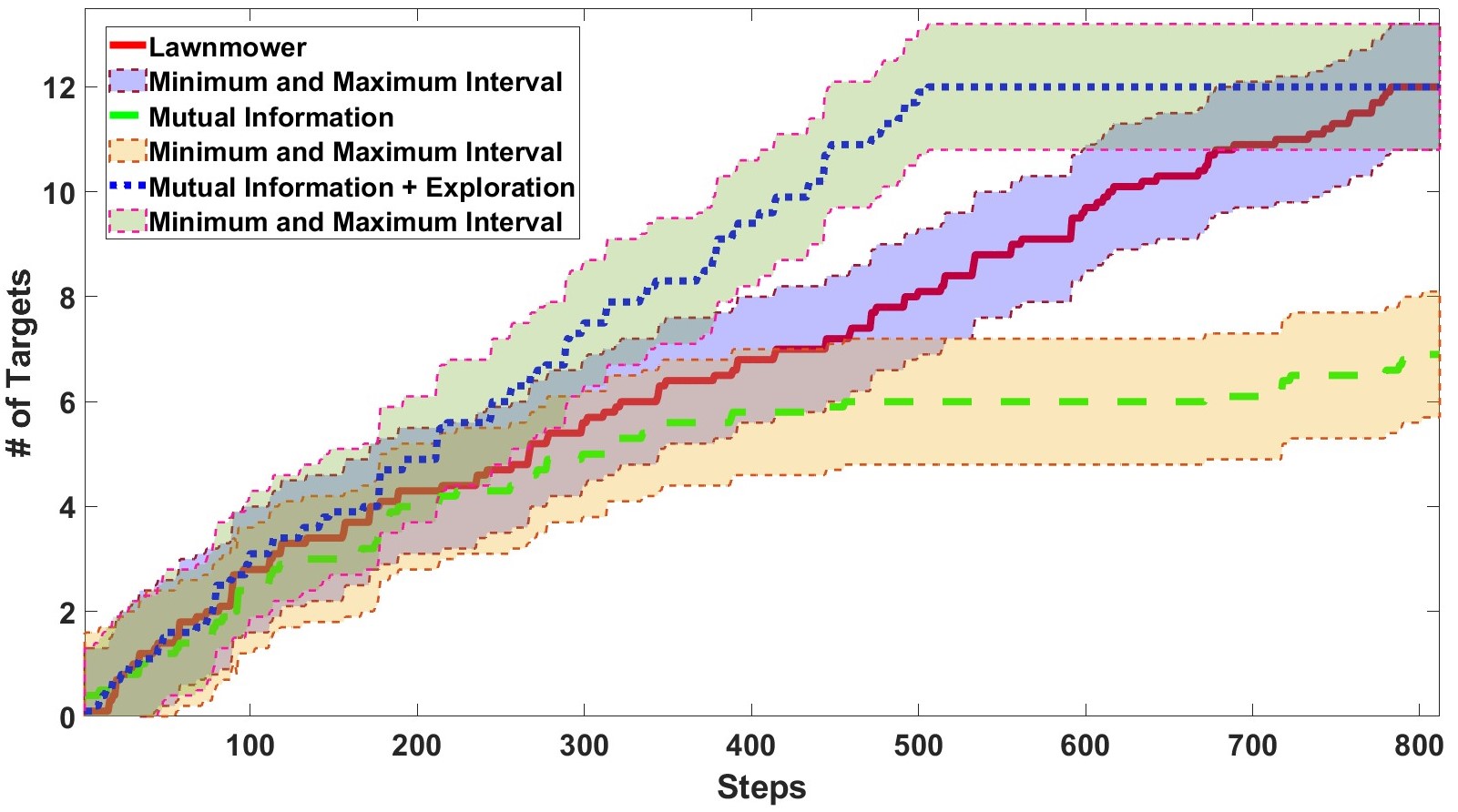}
  \includegraphics[scale=0.21]{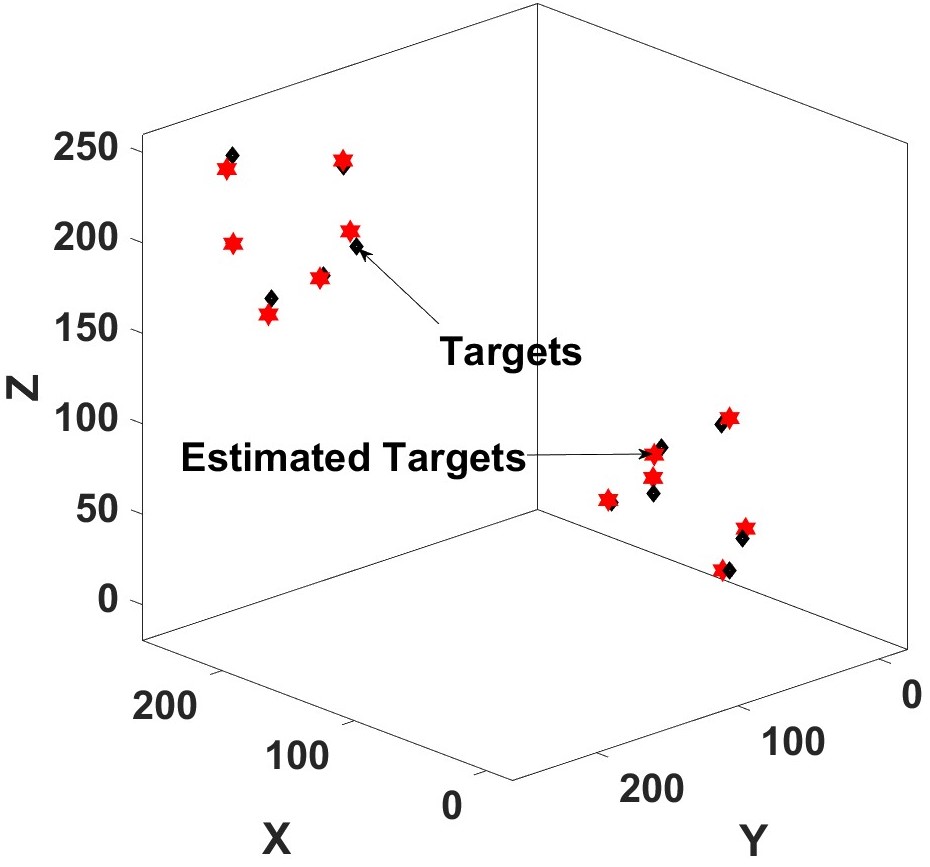}
  \caption{Left: Detected average number of targets placed manually in 2 clusters in 10 random maps. Right: Example of estimation error using our method.}
  \label{Fig.scene10}
\end{figure}
The trajectory length is chosen as $812$ steps since the lawnmower needs this length to complete the search of the whole space. The length is the same for all algorithms for fairness. The results in Fig.~\ref{Fig.scene2} (left) show the number of target detections over time, while Fig.~\ref{Fig.scene2} (right) shows all the algorithms' trajectories in one of the 10 experiments. It is visible that the lawnmower covers the 3D environment uniformly. In contrast, our algorithm focuses on relevant regions to find targets faster. Unlike the MI-only method of \citet{Dam:2015}, our algorithm finds all the targets, thanks to the inclusion of the exploration component.

\medskip \noindent \textit{E3: Planner performance for clustered targets.}~We consider $12$ targets manually placed in $2$ clusters of 6 targets, each at random locations. The trajectory length is the same as for $E2$. Fig.~\ref{Fig.scene10} (left) shows the number of targets detected over time. We see that the performance of our algorithm is again better than the MI-only method and lawnmower. Fig. \ref{Fig.scene10} (right) shows the position of actual targets as well as estimated target locations respectively. The error between the actual and estimated target (root mean squared error, RMSE,) is 3.14m, relatively small for a domain of size $280^{3}$m$^{3}$. This RMSE value depends on the covariance of the Gaussian noise in the sensor model \eqref{eq:sensor_model}, and the threshold values in Algorithm \ref{algo:horizon}. For instance, we can reduce the errors by making the cluster width threshold smaller, as we show in the next experiment.

\medskip \noindent \textit{E4: Threshold value versus RMSE.}~For 12 targets uniformly distributed at random locations, we used $20$ different values of the cluster radius threshold $\mathcal{T}_{r}$, varying in a range from $0.5$ to $2.4$.
\begin{figure}[t]
    \centering
    \includegraphics[scale=0.15]{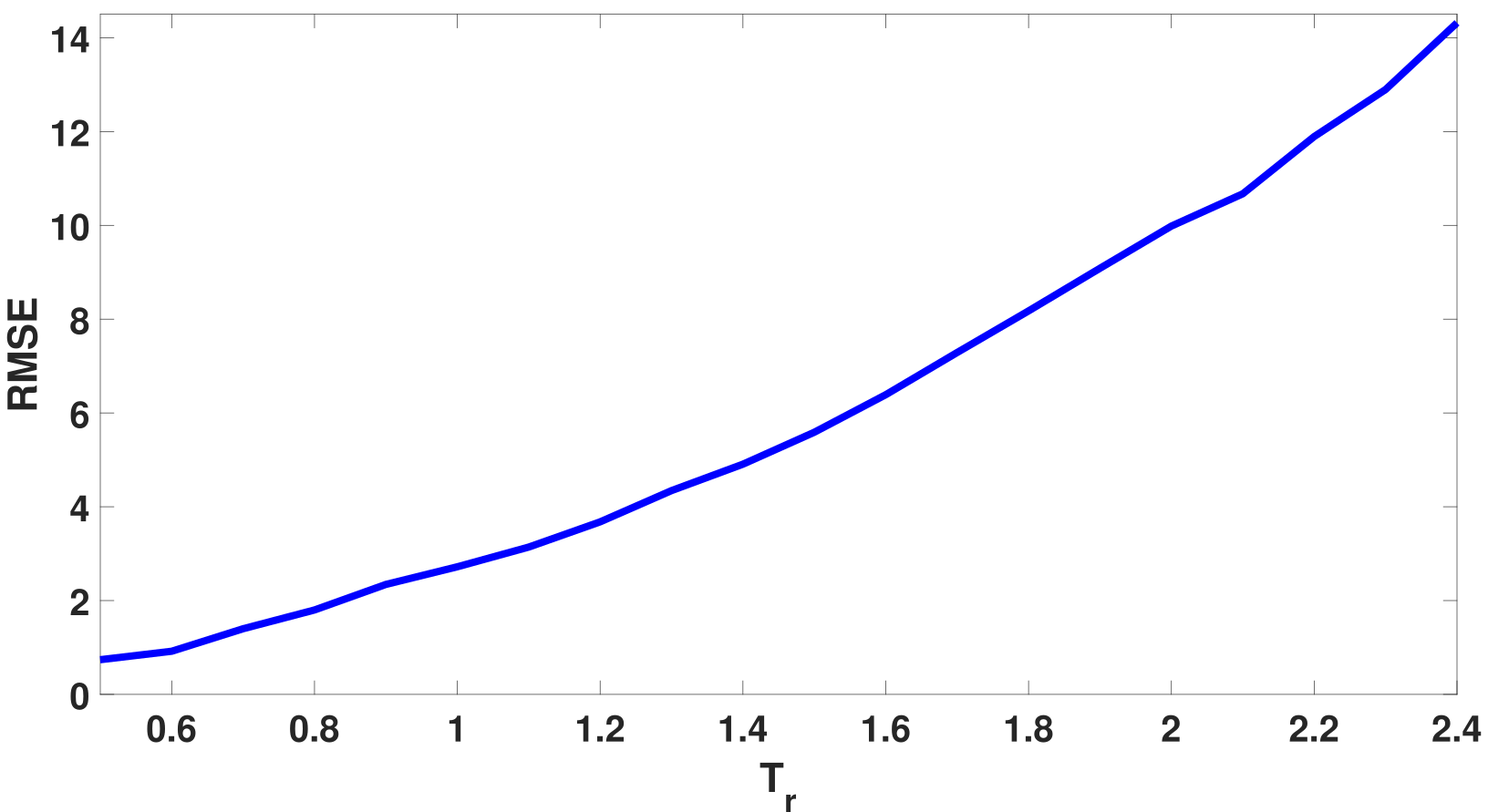}
    \includegraphics[scale=0.15]{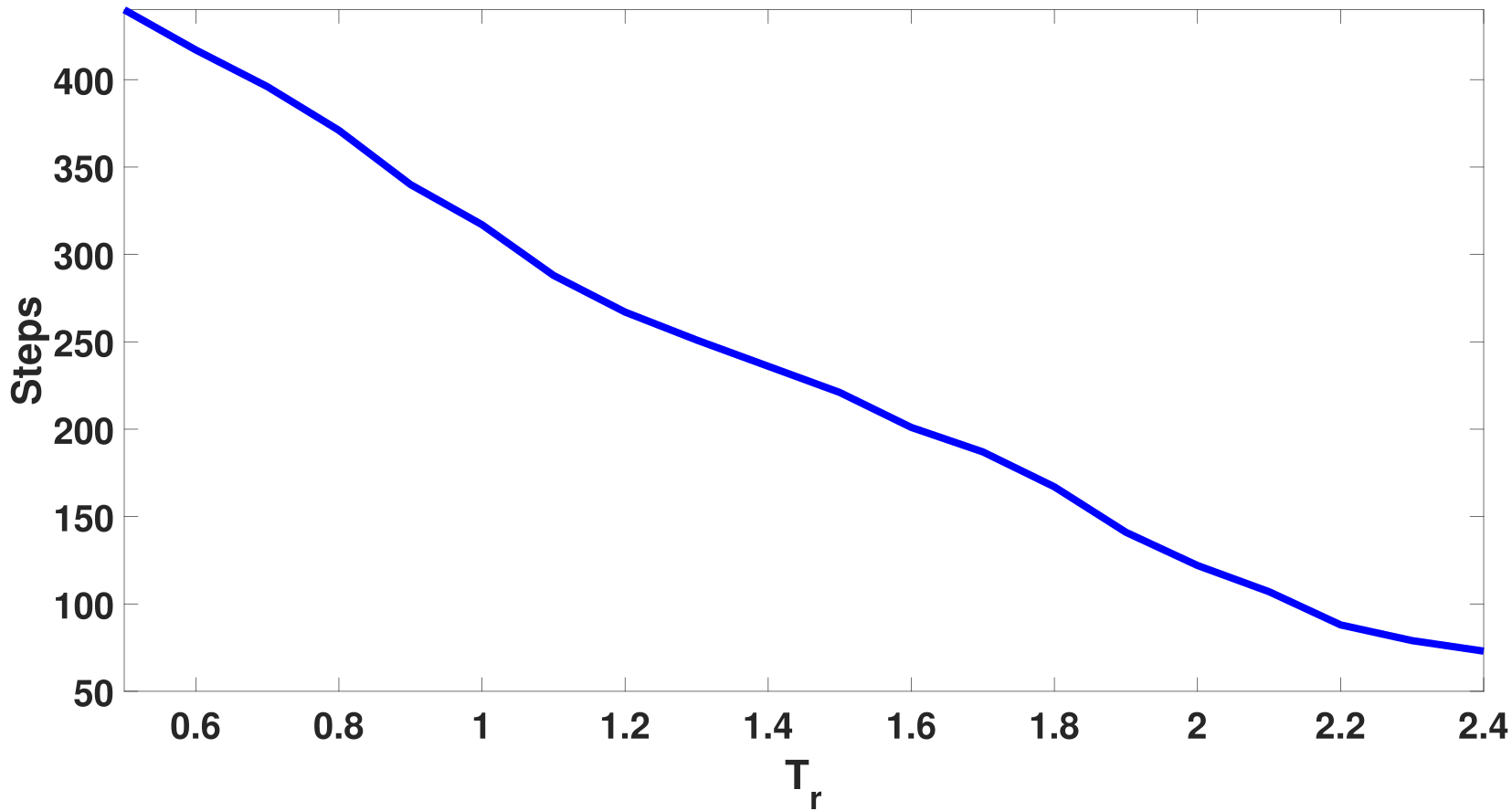}
    \caption{Results from one experiment per values of $\mathcal{T}_{r}$. Left: Target position error for different radius thresholds. Right: Number of steps taken by the UAV to find all targets for the same thresholds.}
    \label{Fig.scene4}
\end{figure}
\begin{figure}[t]
    \centering
    \includegraphics[scale=0.15]{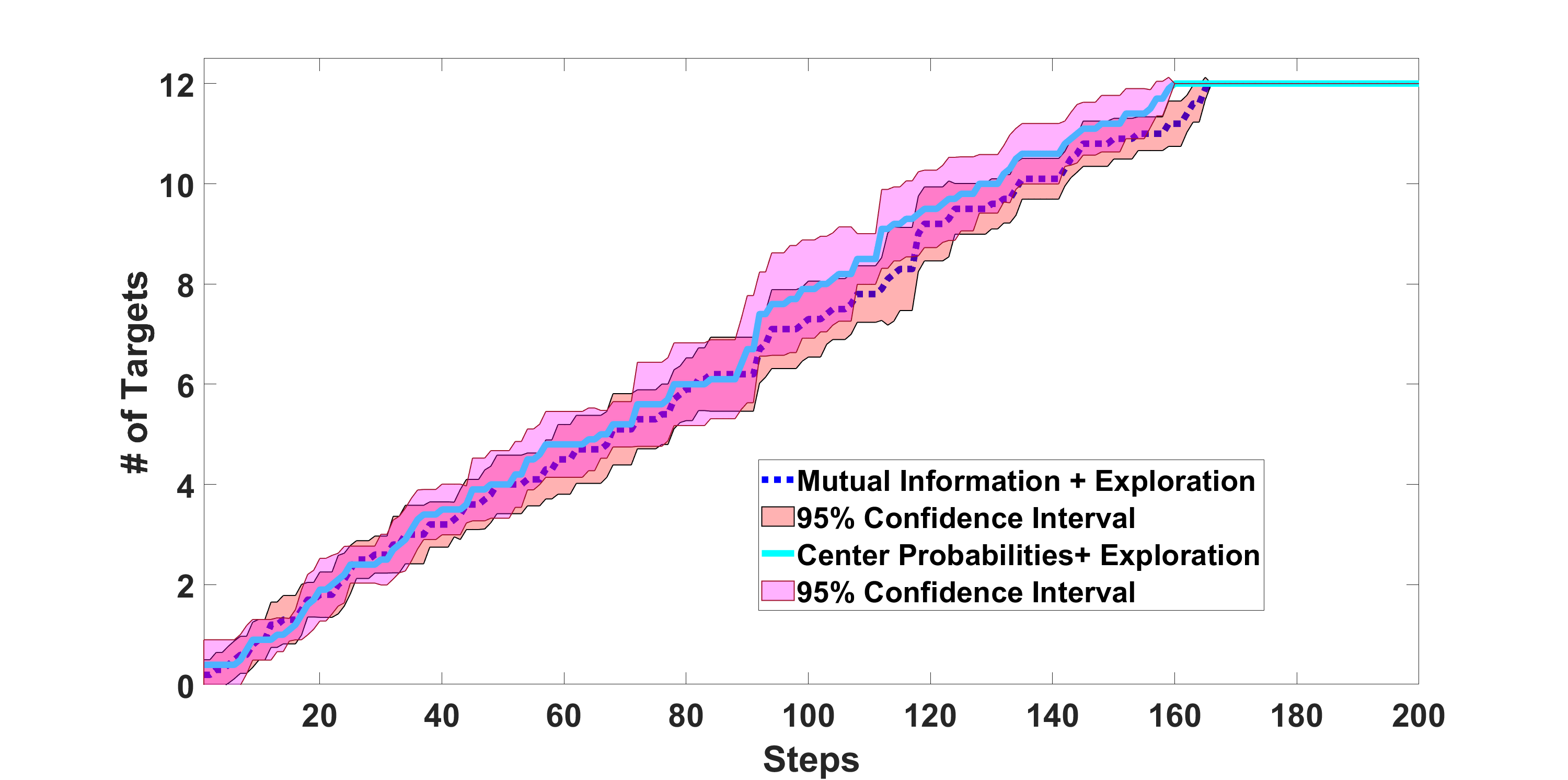}
    \caption{Comparison of detected average number of targets between mutual-information and center-probabilities method in 10 random maps.}
    \label{Fig.scene6}
\end{figure}
\begin{figure}[t]
    \centering
    \includegraphics[scale=0.07]{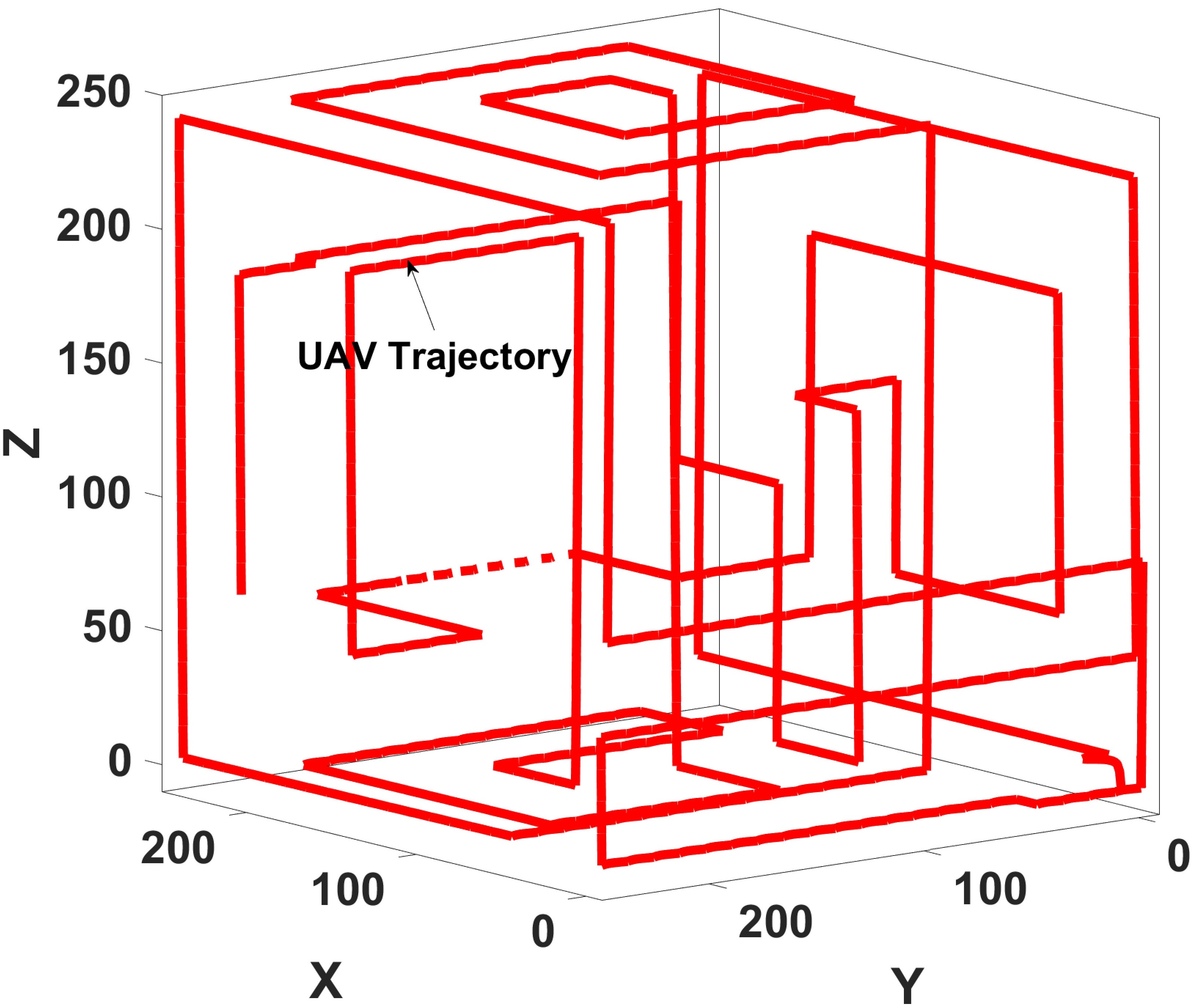}
    \includegraphics[scale=0.07]{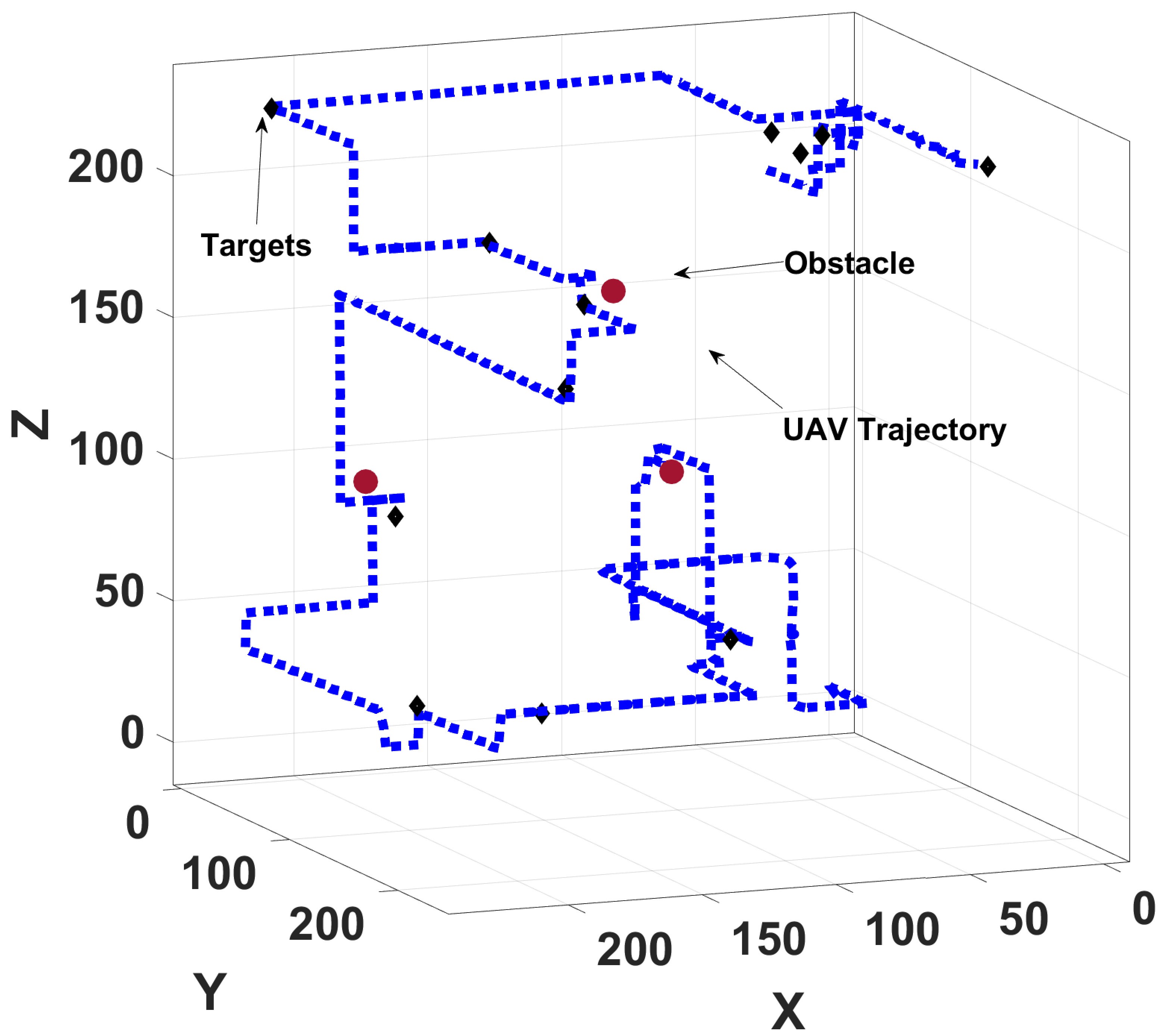}
    \caption{Left: UAV trajectory when there is no target. Right: Illustration of UAV avoiding obstacles while searching for unknown target locations. }
    \label{Fig.scene7}
\end{figure}
The result in Fig. \ref{Fig.scene4} (left) shows the RMSE between the actual and estimated targets locations as a function of the threshold value $\mathcal{T}_{r}$. Errors are directly related to threshold values and can be made smaller by reducing $\mathcal{T}_{r}$. Doing this of course increases the number of steps taken by UAV to find all the targets, as shown in Fig. \ref{Fig.scene4} (right).

\medskip \noindent \textit{E5: Target refinement with center probabilities.}~In this experiment, we study the performance when the target-refinement component of the objective is either MI or the center-probabilities version \eqref{eq:co}. Exploration is always included. We consider 12 targets uniformly randomly distributed. The trajectory length is $200$ steps. Fig.~\ref{Fig.scene6} shows the number of targets detected over time. The algorithm found all targets in a nearly equal amount of steps with the two options of target refinement. The main difference is in computational time: the MI-based algorithm takes an average of 1.11s per step, while center probabilities are faster, with 0.88s per step on average.

\medskip \noindent \textit{E6: Trajectory with no targets.}~We show in Fig.~\ref{Fig.scene7} (left) how the UAV explores the environment in the absence of targets. The trajectory length is chosen as $600$ steps. The UAV flies similarly to a lawnmower pattern because in the absence of target measurements, the exploration component drives the drone to cover the whole environment.

\medskip \noindent \textit{E7: Obstacle avoidance.}~We consider $12$ targets and $3$ obstacles placed manually at arbitrary locations, as shown in Fig.~\ref{Fig.scene7} (right). The trajectory length is $200$ steps. The parameters required for the obstacle avoidance of \eqref{eq:obs_detect}, and \eqref{eq:obs_avoid} are selected to be the danger limit $d_{l}$ is set up to 6 m, and $\mathcal{K}_{obs}=5$.  Fig.~\ref{Fig.scene7} (right) shows the UAV searching for the targets while avoiding obstacles. Due to the obstacles, it takes about 200 steps to find all the targets compared to 170 steps without obstacles.

\medskip
Next, we present our hardware setup and experimental results.

% -------------------------------------------------------------------------
% ----- SECTION BREAK -------------------------------
% ------------------------------
\section{Experimental Results}\label{sec:experiment}
In this section, real-life implementation results are given using a Parrot Mambo drone. Since creating real free-floating static targets would be overly difficult, for this experiment the problem is reduced to 2D search, with the targets on the floor and the drone flying at a constant altitude. % The MATLAB Simulink model provided by Mathworks is used as a baseline in our research, and modified according to our requirements.

\begin{figure}[t]
  \centering
  \includegraphics[scale=0.25]{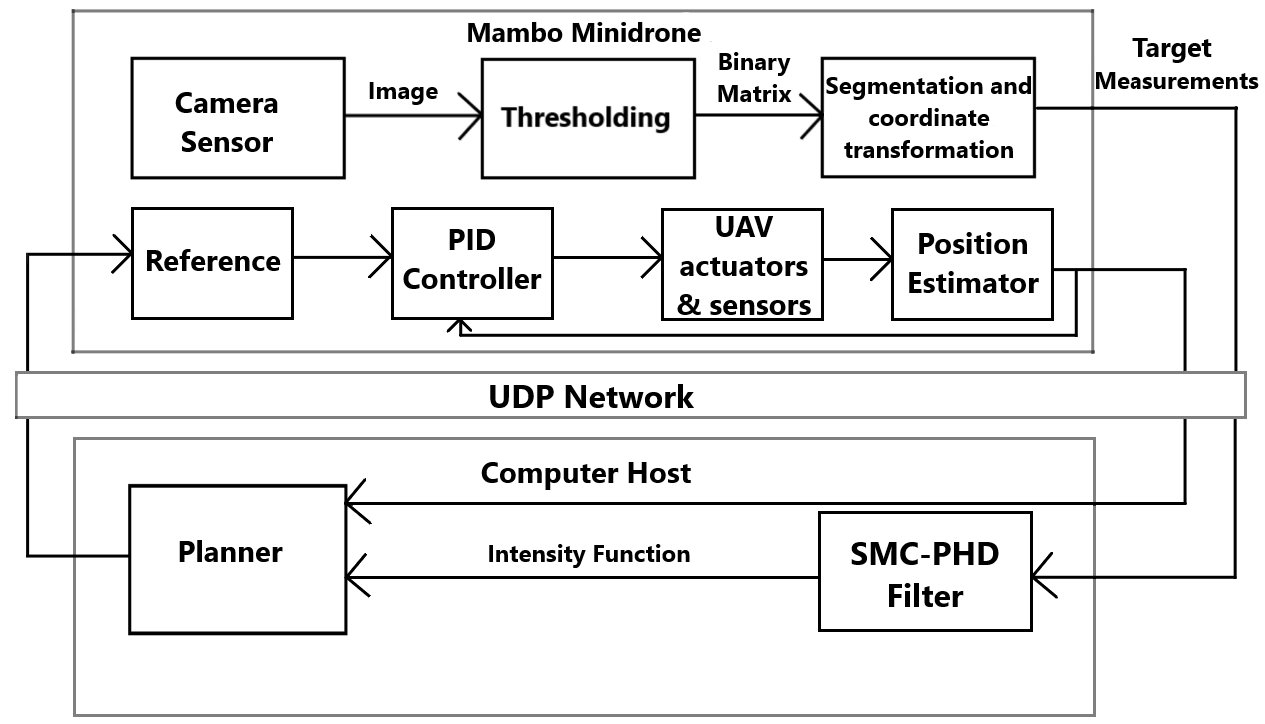}
  \caption{Block diagram of the Parrot Mambo application.}
  \label{Fig.model}
\end{figure}

We start with an overview of the experimental framework, in Fig.~\ref{Fig.model}. After getting the measurements, we update the target intensity function and generate with the planner a new reference position for the drone to move. However, the Parrot Mambo minidrone has little memory and computation power, so it is not possible to do all the processing onboard.~To solve this problem, we created two different nodes, one deployed on the drone, and the second used remotely on the computer.~To share information between these nodes, we establish a UDP communication channel.~On the minidrone, we implement low-level segmentation of the images and perform a coordinate transformation to get the measurements.~Then we transmit those measurements to the host model via UDP. On the computer host we run particle filtering and the planner algorithm to generate the next desired position at each step.~After we get the new reference position generated by the planner, we transmit it to the minidrone via UDP. This reference is enforced using the controller of the minidrone.

Next, we describe the hardware and low-level software in Section \ref{subsec:low}. The high-level setup for the experiment, as well as the real-world results illustrating the performance of our algorithm are described in Section \ref{subsec:setup}.

\subsection{Hardware, Sensing, and Control}\label{subsec:low}

\begin{figure}[!t]
  \centering
  \includegraphics[scale=0.25]{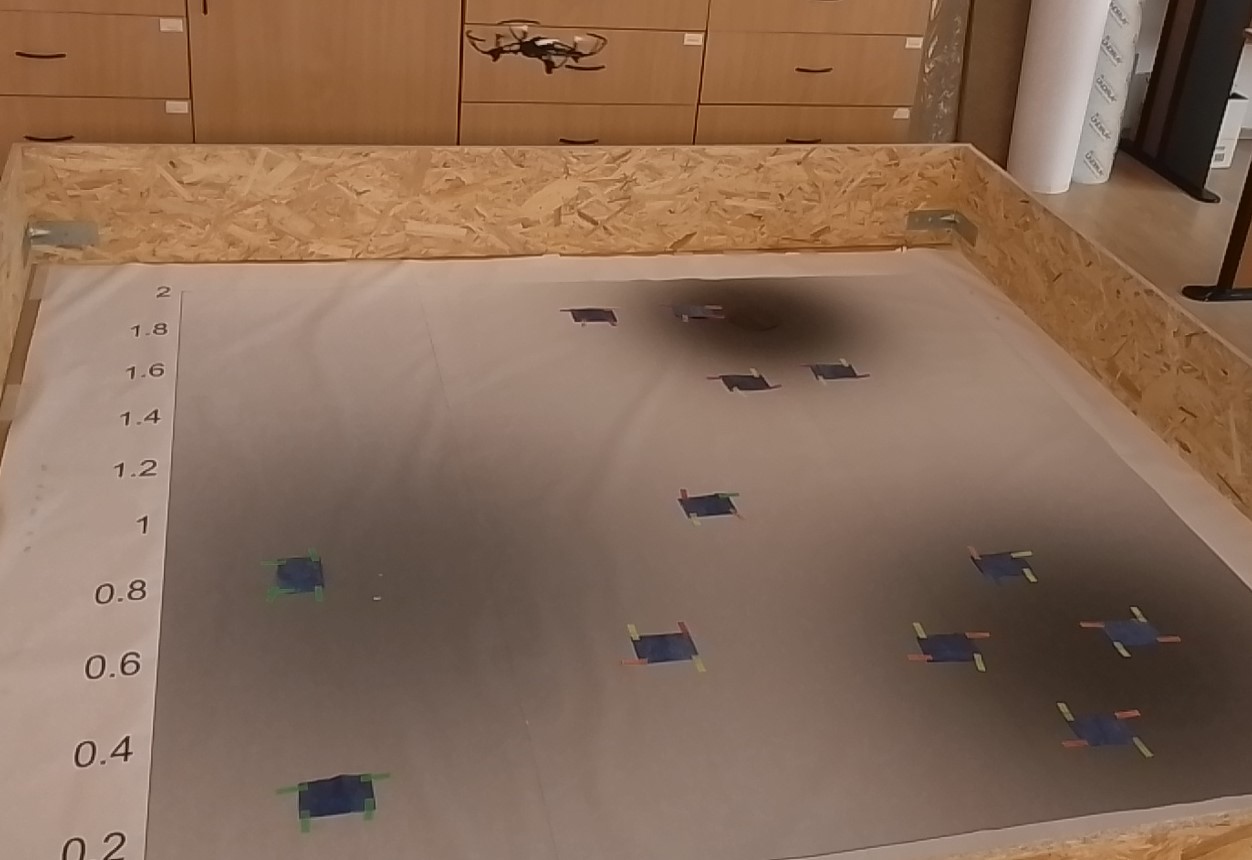}
  \caption{Parrot Mambo minidrone searching for targets (blue markers).}
  \label{Fig.snapshot}
\end{figure}

The Parrot Mambo minidrone is a quadcopter that can be commanded through MATLAB via Bluetooth.~Fig~\ref{Fig.snapshot} illustrates the drone searching for unknown targets. The drone is equipped with an Inertial Measurement Unit (IMU) containing a 3-axis accelerometer and a 3-axis gyroscope, ultrasound and pressure sensors for altitude measurements, and a downward-facing camera that has a 120x160 pixel resolution at 60 frames per second, useful for optical flow estimation and image processing \cite{Maer:2020}. The Matlab/Simulink support package \cite{Kus:2021} allows accessing the internal sensor data and deploying control algorithms and sensor fusion methods in real-time.

%%Thanks to the “cut-out” system, which activates in case of impact, and crash, it shuts down the motor of the drone.

We placed several markers that represent targets on the ground.~Through the use of the drone-mounted camera, pictures containing these markers are taken and fed into an image processing algorithm for the identification and localization of the markers.~The field of view of the camera depends on the flight altitude. Maintaining a proper altitude is crucial \cite{Kus:2021}, as flying at a lower altitude than the prescribed one will often place the markers outside the image frame, while flying higher can make the markers too small and difficult to distinguish.

\begin{figure}[t]
\centering
   \includegraphics[scale=0.231]{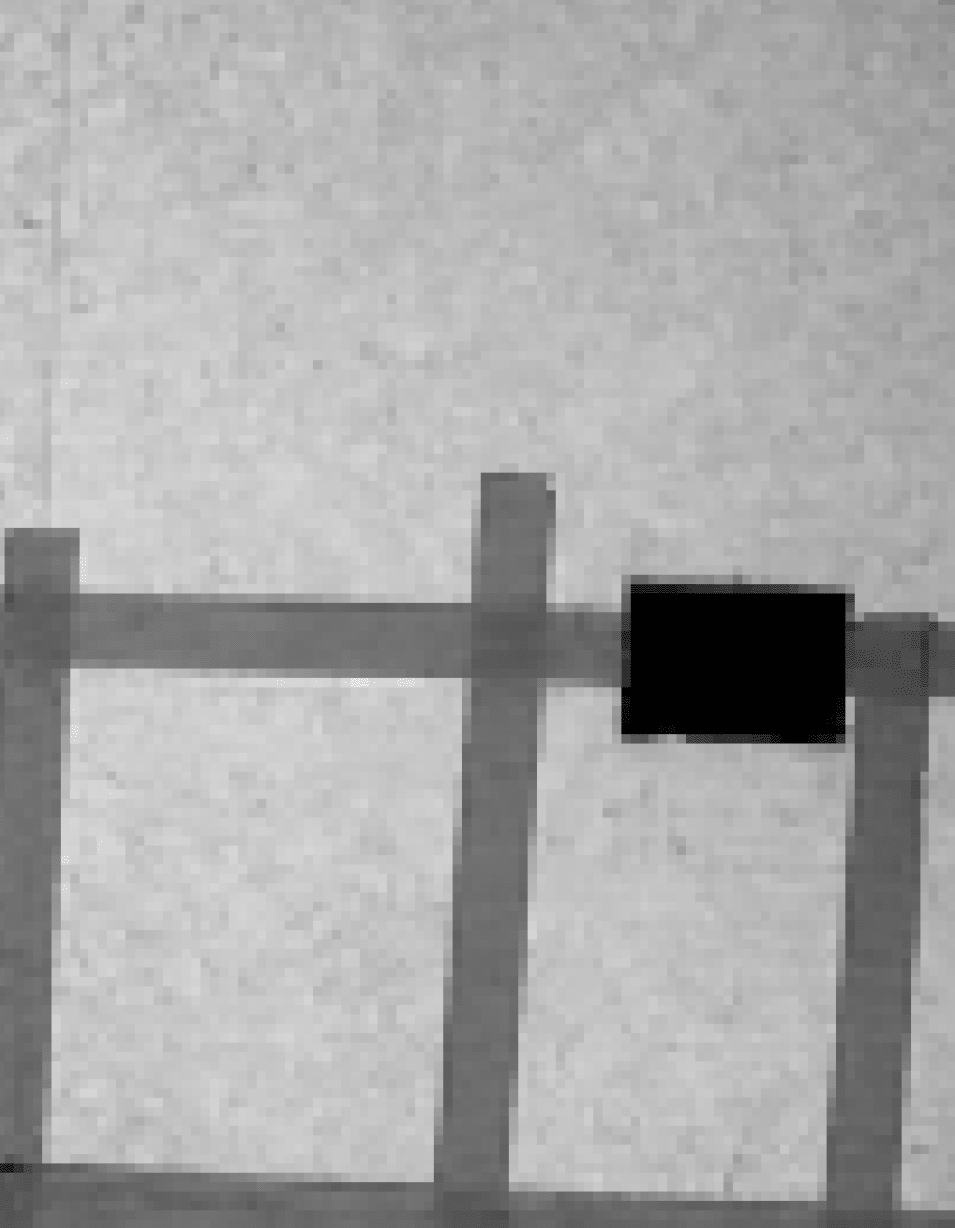}
   \includegraphics[scale=0.26]{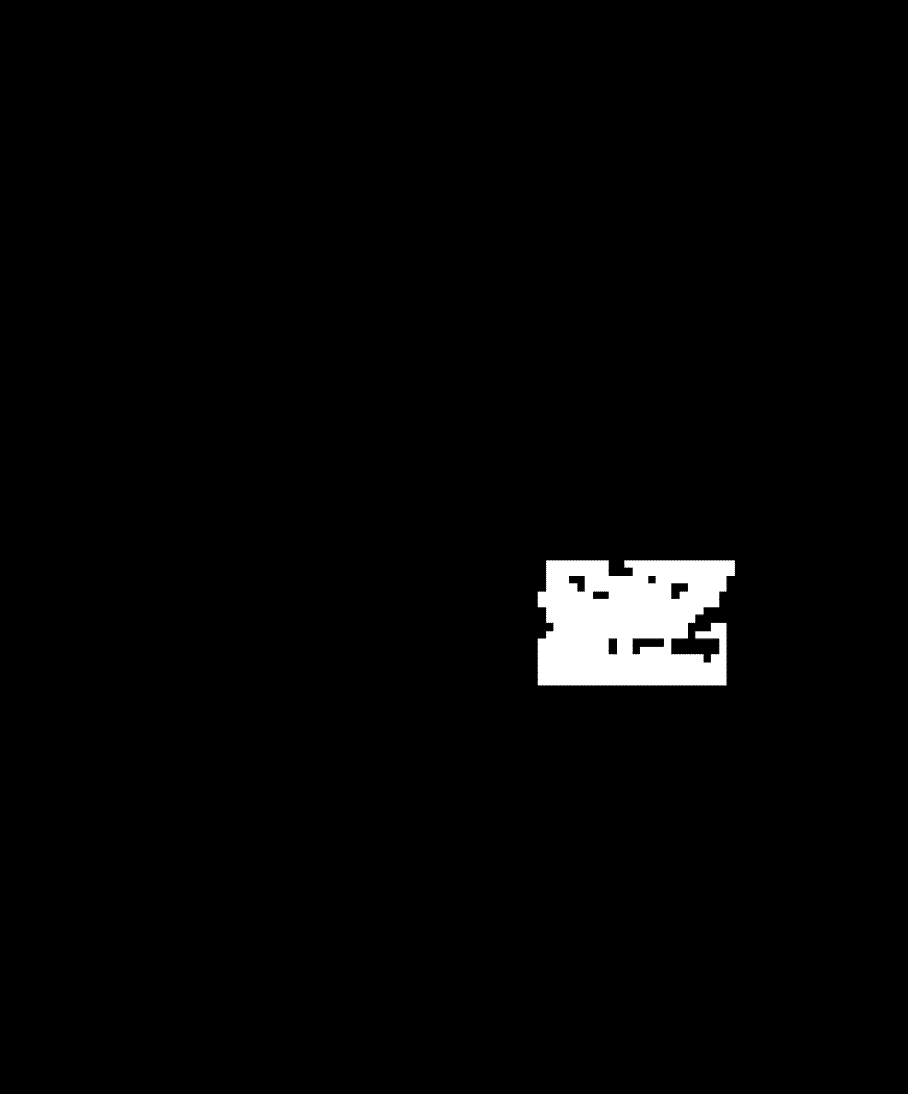}
    \caption{Example of blue color thresholding Left: before thresholding. Right: after thresholding}
    \label{Fig.threshold}
\end{figure}
Given an RGB image $\mathbb{G}=(R,G,B)$  where $R,G,B$ are each an $n\times m$ dimensional matrix, image segmentation was used to identify the blue-colored markers in the images. One simple way of doing this is through thresholding \cite{Mil:2008}. First, we find the pixels that have blue as the dominant color, by computing matrix $ O_{p} = B - \frac{R}{2} - \frac{G}{2}$, where operations are applied element wise. After applying a single level threshold of value $\mathcal{T}_{im}$, the resulting binary image $\mathbb{T}_{im}\in \left\{0,1\right\}^{n\times m}$ is:
\begin{equation}
\begin{aligned}
\mathbb{T}_{im}(x_{im},y_{im})=\begin{cases} 0,& O_{p}(x_{im},y_{im})<\mathcal{T}_{im}\\
1,& O_{p}(x_{im},y_{im})>\mathcal{T}_{im} \end{cases}
\end{aligned}
\end{equation}
% The results can vary substantially depending on lighting conditions.
An example of thresholding is shown in Fig.~\ref{Fig.threshold}, with the threshold $ \mathcal{T}_{im}= 35$.

After segmentation, we compute the marker position relative to the world reference frame. A backward projection method is performed on the marker centroid position in the image. Define $[X_{c},Y_{c},Z_{c}] \thinspace,\thinspace [X_{w},Y_{w},Z_{w}]$  to be the camera and world reference frames, respectively. The image plane coordinates of the projection $P_{im}$ of a centroid point are represented in homogeneous world coordinates as $P_{h}= \begin{bmatrix}
x_{h} &y_{h} & z_{h} & 1
\end{bmatrix}^{T}$.
Here, the targets (markers in this case) are on the floor, so ${Z}_{w}$ is taken zero. Then, $P_{h}$ can be obtained by taking the inverse~ of:
\begin{equation}\label{eq:world}
\begin{aligned}
P_{im}=\begin{bmatrix}x_{im} & y_{im} & 1\end{bmatrix}^{T} = \Upsilon_{im} \Lambda P_{h}
\end{aligned}
\end{equation}
where $\Lambda$ defines the transformation matrix between the word and camera reference frame and $\Upsilon_{im}$ is the camera intrinsic matrix  \cite{Maer:2020}. The transformation matrix is a 4×4 rotation matrix based on the position of a UAV $(x_{q}, y_{q}, z_{q})$ centered on the field of view. It is used to transform the target coordinates from camera frame to the world reference frame. For a CCD camera, the intrinsic matrix has the general form:
\begin{equation}\label{eq:inter}
\Upsilon_{im} =\begin{bmatrix}
\ss_{x} &0
&x_{ic}  &0 \\
0 &\ss_{y}
&y_{ic}  &0 \\
0 & 0 & 1 &0
\end{bmatrix}
\end{equation}
where $\ss_{x}$ and $ \ss_{y}$ represent the pixel focal length and camera FOV, and are identical for square pixels. Moreover, $(x_{ic}, y_{ic})$  are the image plane coordinates expressed in pixels. To get the values of  $\ss_{x},\ss_{y}$, we use the image dimensions and the camera angle in the horizontal and vertical directions. The change of reference frame gives a pair of coordinates that represent a target measurement.

For low-level control, instead of the 3D backstepping controller from the simulations (Section \ref{subsec:control}), we use the built-in altitude and position controllers from the Simulink model provided by Mathworks \cite{Kus:2021, Abd:2019} to move the drone in 2D. These controllers are simpler and their performance is sufficient for our problem. In practice, the continuous-time controller is implemented  using time discretization with a low-level sampling period of $0.005$s.

\subsection{High-Level Setup and Results}\label{subsec:setup}

% 2D case for: h, then just a sentence saying things go thru for 2D variables in filtering, the planner, and target removal, with the note that interpolation is bilinear on a 2D grid.

Our indoor experiments run in a 2D environment having a size of $[-0.3, 2]$m$\thinspace\thinspace\times\thinspace[-0.3, 2]$ m, in which the Parrot Mambo minidrone searches for the unknown targets (blue markers in this case). In this experiment, the action set has eight different candidates: forward, backward, right, left, and the diagonals. The drone moves $0.2$m at each step and flies at a constant altitude of 1m above the ground to maintain the same size of the field of view.

 %The MATLAB/Simulink is used to deploy the framework to the parrot mambo minidrone, and it is also used for the communication from host to minidrone model and vice versa.%

Due to the particularities of the hardware experiment, the sensor model is different from the one of Section \ref{sec:problem}. The detection probability is:
\begin{equation}\label{pd_detection}
\pi(x,q)=\begin{cases}
 1 &  x_{ik} \thinspace \in\thinspace \mathcal{F}  \\
0 &  x_{ik} \thinspace \notin\thinspace \mathcal{F}
\end{cases}
\end{equation}
where $\mathcal{F}=[\mathcal{X}_{q}-0.2,\mathcal{X}_{q}+0.2]\thinspace\thinspace\times\thinspace[\mathcal{Y}_{q}-0.2,\mathcal{Y}_{q}+0.2]$ is the camera's field of view at the 1m altitude of the UAV. Target measurements have a form similar to \eqref{eq:sensor_model}:
\begin{equation}\label{eq:2Dsensor_model}
Z_{k}=\bigcup_{i\thinspace\in\left\{1,\hdots,N_{k}\right\}\thinspace \mathrm{s.t}. \thinspace b_{ik}=1}\left[\hat{h}_{ik}(x_{ik})+\hat{\varrho}_{ik}\right]
\end{equation}
but now, $\hat{h}_{ik} = [d_{ik}, \theta_{ik}]^{T}$ contains only the Euclidean distance $d_{ik}$ between the 2D positions $q_{k}$ and $x_{ik}$ and the bearing angle $\theta_{ik}$, computed as in \eqref{eq:sensor_model}. Moreover, $\hat{\varrho}_{ik}$ is 2D Gaussian noise. To estimate its covariance $R$, we captured 100 images of a single target from different positions of the drone, and then computed the empirical covariance of these measurements. We obtained $R=\mathrm{diag}[0.145, 0.112]$. Note that due to the binary $\pi$, $b_{ik}=1$ if and only if $x_{ik} \in \mathcal{F}$. 

% To calculate the target measurement density $ g(z_{k}|x_{k})=\mathcal{N}(z_{k},h(x_{k}),R)$, 

The filter and planner equations of Sections \ref{sec:filter} and \ref{sec:exploration} hold in 2D, with the mention that the exploration bonus $\iota$ is now interpolated bilinearly on a 2D grid.

To validate our algorithm, we initialized the drone at the $[0;0;0]$ coordinates. We set the maximum number of particles to 5000 to reduce the computation time. The threshold values in Algorithm \ref{algo:horizon} are set experimentally at $\mathcal{T}_{r}=0.1$,$\mathcal{T}_{m}=0.2$, and $\mathcal{T}_{z}=0.3$. Up to $100$s are allocated to complete the experiment. The high-level sampling period used for estimation, generating new waypoints, and communication is $\Delta_{H}=3.05$s.

 \begin{figure}[t]
  \centering
   \includegraphics[scale=0.2]{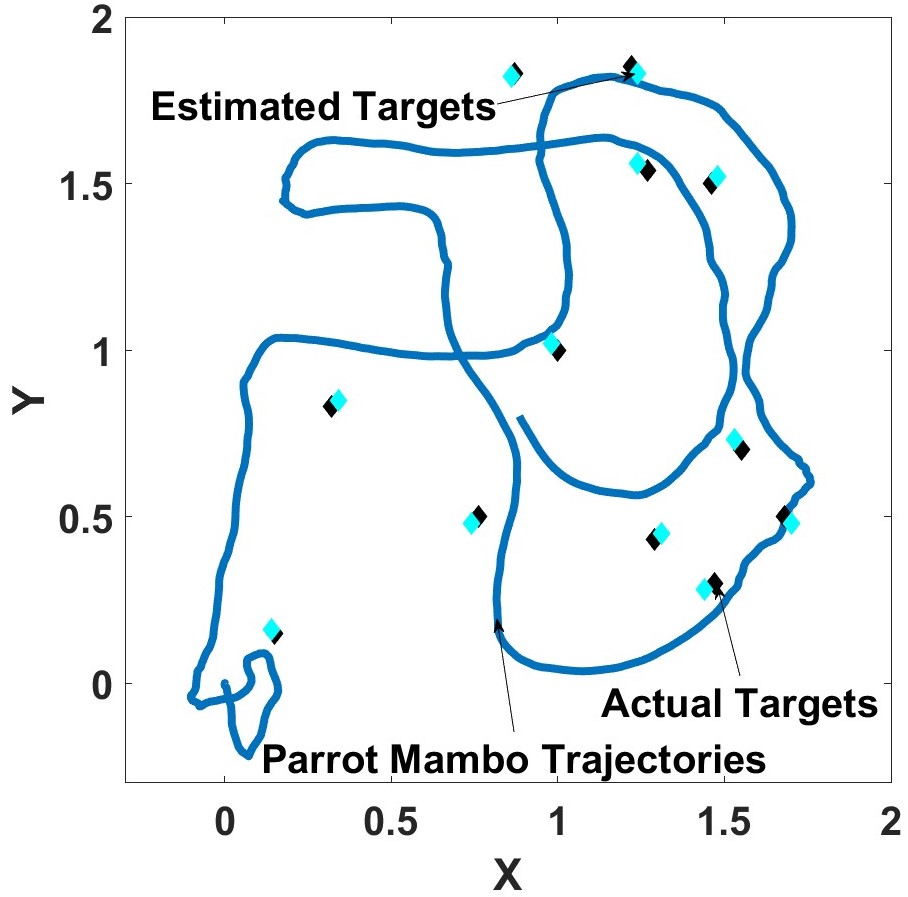}
   \includegraphics[scale=0.12]{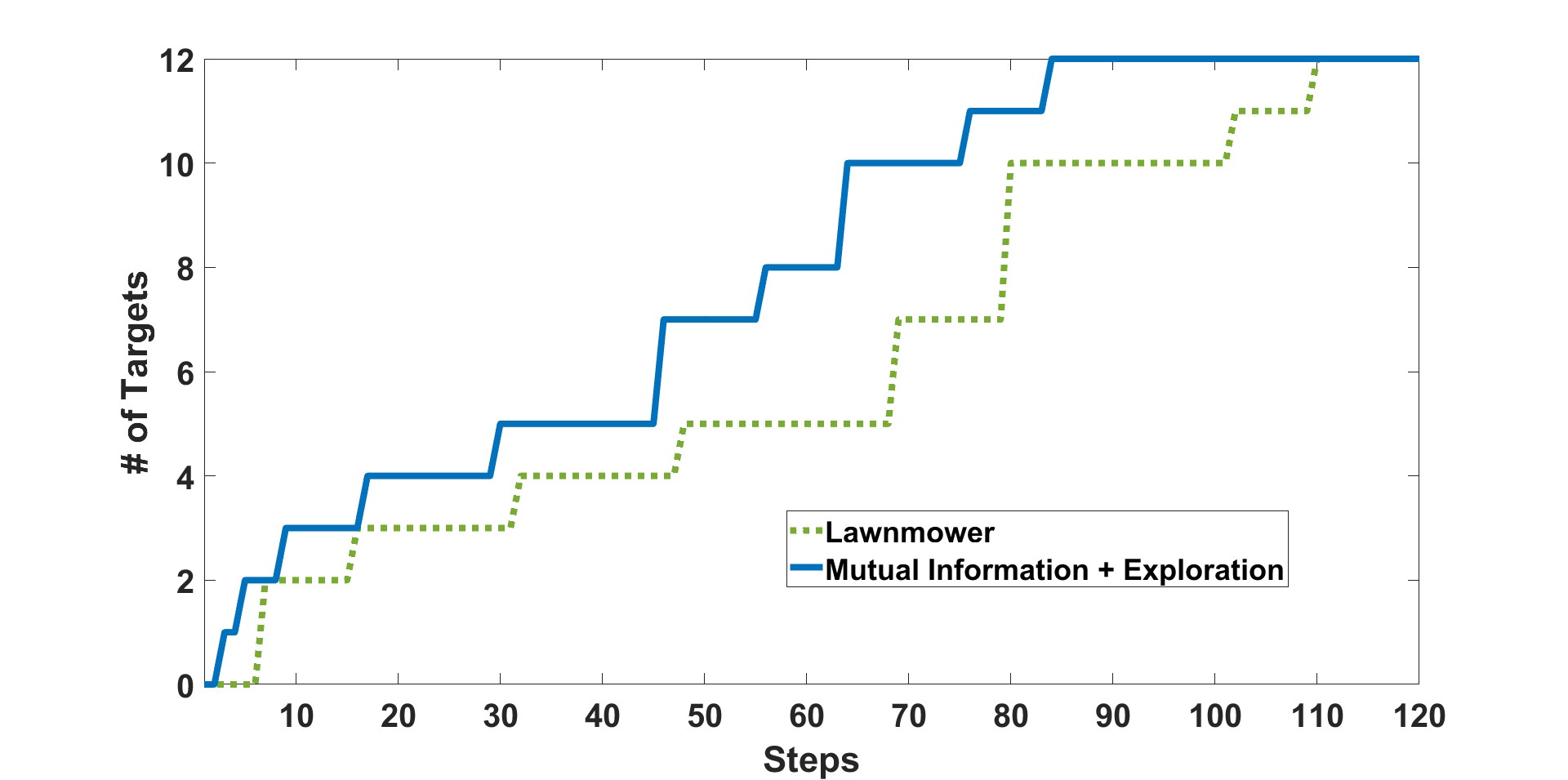}
  \caption{Left: Trajectories with our algorithm on the 2D real-world with actual and estimated targets. Right: Detected number of targets using real Parrot Mambo minidrone}
  \label{Fig.exp_result}
\end{figure}
We consider $12$ targets manually placed at arbitrary locations. Fig.~\ref{Fig.exp_result} (left) shows our algorithm's trajectory, together with the actual and estimated target locations. The drone finds the targets, with an RMSE between the actual and estimated target locations of 0.08 m, which is relatively small. There is some drift because we rely on the onboard sensors. A video of a real-world experiment is available online at \url{http://rocon.utcluj.ro/files/mambo_targetsearch.mp4}.

Fig.~\ref{Fig.exp_result} (right) shows the number of target detections, compared to a real-life implementation of the lawnmower. The $X-Y$ lawnmower spacing in each such layer is set to $0.2$ meters. Like in the simulations, the proposed method works better than the lawnmower.

\section{Conclusions}\label{sec:conclusion}
In this paper, we addressed the problem of finding an unknown number of static targets using a UAV. The proposed method is composed of a multi-target SMC-PHD estimator, a planner based on exploration and target refinement, a method to remove found targets, and a backstepping nonlinear control algorithm with obstacle avoidance. The effectiveness of the proposed algorithm was validated through extensive simulations and in a real-robot experiment. The results show that the algorithm works well both for uniformly distributed targets and sparse target clusters; as well as for a real drone.

In future work, we aim to include control cost in the planning objective. Another possibility for the Mambo experiment is to use optical tracking to eliminate drift and get a more accurate position of the drone.

\backmatter

\bmhead{Supplementary information}
Not applicable to this article.

\bmhead{Acknowledgments}
The authors gratefully acknowledge the funding sources described below.

\bmhead{Author Contribution}
All authors contributed to the design and evaluation of the method and experiments, as well as with feedback on manuscript versions. Implementation, simulation and hardware experiments were performed by Bilal Yousuf. The paper was written by Bilal Yousuf and Lucian Bu\c{s}oniu. Zs\'{o}fia Lendek helped in designing the low-level controller. All authors read and approved the final manuscript.

\bmhead{Funding}

This work was been financially supported from H2020 SeaClear, a project that received funding from the European Union’s Horizon 2020 research and innovation programme under grant agreement No 871295; and by the Romanian National Authority for Scientific Research, CNCS-UEFISCDI, SeaClear support project number PNIII-P3-3.6-H2020-2020-0060 and Young Teams project number PN-III-P1-1.1-TE-2019-1956.
\section*{Declarations}

\bmhead{Conflict of interest} The authors have no competing interests to declare that are relevant to the content of this article.

\bmhead{Consent to participate} Not applicable (this is not a study with human participants).

\bmhead{Consent for publication} Not applicable (this is not a study with human participants).

\bmhead{Availability of data and materials} Not applicable.

\bmhead{Code availability} Not applicable.

\bmhead{Ethics approval} Not applicable.


\noindent


\begin{appendices}

\section{Multitarget filtering}\label{sec:appendix}
In this appendix, we describe in more detail the Sequential Monte Carlo based Probability Hypothesis Density (SMC-PHD) filter. This framework is adopted from \cite{Dou:2005}.~We start with the random finite set (RFS) formulation, followed by discussing the intensity function.~Later on, we define the Sequential Monte Carlo-based approximation method, followed by resizing and resampling the set of particles.

\subsection*{Random Finite Set Model}
In multi-target search, the states and observations are two \emph{collections} of individual targets and measurements.~The set of targets $X_{k}=\left \{ x_{1k},x_{2k},....,x_{N_{k}k} \right \}\thinspace\subset\thinspace E$ is the realization of the target RFS $\Xi_{k}$ \cite{Dou:2005} at step $k$.~The set of measurements $Z_{k}=\left\{ z_{1k},z_{2k},\hdots,z_{M_{k}k}\right\}$ is similarly a realization of the measurement RFS $\Sigma_{k}$. The multi-target state and measurement RFSs $\Xi_{k}$ and $\Sigma_{k}$, obey:
\begin{equation}\label{eq:RFS_model}
\begin{aligned}
\Xi_{k}=&s_{k}(X_{k-1})\cup \Gamma_{k}\\
\Sigma_{k}=&\Theta_{k}(X_{k})
\end{aligned}
\end{equation}
Here, $\Xi_{k}$ comprises the multi-target evolution, in which $s_{k}(X_{k-1})$ denotes the surviving targets at time step $k$ depending on the previous set of targets $X_{k-1}$, and $\Gamma_{k}$ defines the target birth term. $\Sigma_{k}$ comprises the sensor measurements $\Theta_{k}(X_{k})$ generated by a set of targets $X_{k}$.

\subsection*{Intensity Function}
The Probability Hypothesis Density (PHD), or intensity function, is a function defined over the target space with the property that its integral over any region $S$ is the expected number of targets in that region. To define it formally, consider first a given set of targets~ $X$. A subset $S\thinspace \subseteq \thinspace E$ contains a number of targets ${\footnotesize N_{X}(S)=\sum_{x\thinspace\in\thinspace X}1_{S}(x)=\left|X \cap S\right|}$. where $1_{S}(x)$ is the indicator function of set $S$. The RFS $\Xi$, of which one realization is $X$, is similarly defined by the \emph{random} counting measure $N_{\Xi}(S)=\left|\Xi \cap S \right|$.~The intensity measure $V_{\Xi}$ is defined as:
\begin{equation*}
\begin{aligned}
&V_{\Xi}(S)= \mathrm{E}[N_{\Xi}(S)]
\end{aligned}
\end{equation*}
for each $S\subseteq E$. $V_{\Xi}$ thus gives the expected number of elements of $\Xi$ in $S$.~The intensity function $D_{\Xi}$ is defined as:
\begin{equation}\label{eq:PHD}
D_{\Xi}=\frac{dV_{\Xi}}{d\lambda}
\end{equation}
where $\lambda$ is the Lebesgue measure \cite{Hun:1997}.~The intensity function is similar to a probability density function, with the key difference that its integral over the entire domain is not $1$, but the number of targets. An example of the intensity measure is given in Fig.~\ref{Fig.particles_filtering}, where the sets $S$ are the squares (this is just an example, they could have any shape) and the shades of gray are the values $V_{\Xi}$ for each $S$.

\subsection*{SMC-PHD Filter}
The overall PHD filter \cite{Dou:2005} was summarized in \eqref{eq:summerized} of Section \ref{sec:filter}. There, $D_{k|k-1}$ is the prior intensity function and $D_{k|k}$ denotes the posterior.~

Let $D_{k-1|k-1}$ represent the intensity function corresponding to the multi-target prior at time step $k-1$, for $k\geq 1$. This intensity function is represented in terms of particles $\hat{D}_{k-1|k-1}(x_{k-1})$ $=\sum_{i=1}^{L_{k-1}}\omega_{k-1}^{i}\delta_{x_{k-1}^{i}}(x_{k-1})$.~This representation reduces the computational complexity of the filter updates. Note that we reuse $i$ for particle index, and that $E\left [ \left | \Xi_{k} \cap S \right | \big{|} Z_{1:k} \right ]\approx$ $\sum_{j=1}^{L_{k}} 1_{s}(x_{k}^{i})\omega_{k}^{j}$: the expected number of targets in a set $S$ is equal to the sum of the weights of the particles in that set, see again Fig.~\ref{Fig.particles_filtering}. Substituting $\hat{D}_{k-1|k-1}$  in \eqref{eq:PHD_predict}, one gets:
\begin{equation}\label{eq:SMC_PHD_predict}
\begin{aligned}
\Phi_{k|k-1}(\hat{D}_{k-1|k-1})&(x_{k})=\sum_{i=1}^{L_{k-1}}\omega_{k-1}^{i}\mathcal{O}_{k|k-1}(x_{k},x_{k-1}^{i})+\Upsilon_{k}(x_{k})
\end{aligned}
\end{equation}
A particle approximation of $\Phi_{k|k-1}(\hat{D}_{k-1|k-1}) (x_{k})$ can be derived by applying importance sampling to each of its terms. For this we need the importance (or proposal) densities $\tilde{p}_{k}(\cdot |Z_{k}),$ and $\tilde{r}_{k}(\cdot |x_{k-1},Z_{k})$. In our specific framework for static target tracking and detection, we take the proposal density for target birth to be $\tilde{p}_{k}(x_{k}|Z_{k})\sim\mathcal{N}(x_{k},\mu(Z_{k}),\thinspace R(Z_{k}))$,  which is a Gaussian defined by the empirical mean $\mu(Z_{k})=\frac{1}{|Z_{k}|}\sum_{z\thinspace\in\thinspace Z_{k}}h^{-1}(z)$ and empirical covariance $R(Z_{k})=\frac{1}{|Z_{k}|}\sum_{z\thinspace\in\thinspace Z_{k}}[h^{-1}(z)-\mu(Z_{k})][h^{-1}(z)-\mu(Z_{k})]^{T}$ of the set of observations $Z_{k}$; where $h$ is the application of the $h_{ik}$ from \eqref{eq:sensor_model}. To calculate the empirical mean and covariance, as the targets are in Cartesian space, we convert the measurement set $Z_{k}$  from polar to Cartesian coordinates by applying $h^{-1}(Z_{k})$. The transition proposal density in the static target scenario can be written as:
\begin{equation*}
\tilde{r}_{k}(x_{k}|x_{k-1}^{i},\thinspace Z_{k})=\delta_{x_{k-1}^{i}}(x_{k}).
\end{equation*}
Equation~\eqref{eq:SMC_PHD_predict} is reformulated as:
\begin{equation}\label{eq:SMC_PHD_predict_weights}
\begin{aligned}
\Phi_{k|k-1} (\hat{D}_{k-1|k-1})(x_{k})=&\sum_{i=1}^{L_{k-1}}\omega_{k-1}^{i}\frac{\mathcal{O}_{k|k-1}(x_{k},x_{k-1}^{i})}{\tilde{r}_{k}(x_{k}|x_{k-1}^{i},Z_{k})} \cdot\\ &\cdot\tilde{r}_{k}(x_{k}|x_{k-1}^{i},Z_{k})+\frac{\Upsilon_{k}(x_{k})}{\tilde{p}_{k}(x_{k}|Z_{k})}\tilde{p}_{k}(x_{k}|Z_{k})\\
\end{aligned}
\end{equation}\label{SMC_PHD_predict_dirac}
Thus, the Monte-Carlo approximation is obtained as:
\begin{equation}
\Phi_{k|k-1}(\hat{D}_{k-1|k-1})(x_{k})=\sum_{i=1}^{L_{k-1}+J_{k}} \omega_{k|k-1}^{i}\delta_{x_{k}^{i}}(x_{k})
\end{equation}
where $L_{k-1}$ is the number of existing particles and $J_{k}$ is the number of new particles arising from the birth process.~Let us denote $L_{k}\equiv L_{k-1}+J_{k}$.~The required particles are drawn according to \cite{Dou:2005}:
\begin{equation*}
\begin{aligned}
x_{k}^{i}&\sim{\footnotesize\begin{cases}
\tilde{r}_{k}(\cdot|x_{k-1}^{i},Z_{k}) & i=1,\hdots,L_{k-1} \\
\tilde{p}_{k}(\cdot|Z_{k}) & i=L_{k-1}+1,\hdots,L_{k}
\end{cases}}
\end{aligned}
\end{equation*}
The weights of the particles are computed as \cite{Dou:2005}:
\begin{equation*}
\omega_{k|k-1}^{i}={\footnotesize\begin{cases}
\frac{\omega_{k-1}^{i}\mathcal{O}_{k|k-1}(x_{k}^{i},x_{k-1}^{i})}{\tilde{r}_{k}(x_{k}^{i}|x_{k-1}^{i},Z_{k})} &  i= 1,\hdots,L_{k-1}\\
\frac{\Upsilon_{k}(x_{k}^{i})}{J_{k}\cdot \tilde{p}_{k}(x_{k}^{i}|Z_{k})}&  i=L_{k-1}+1,\hdots,L_{k}
\end{cases}}
\end{equation*}
Next, we discuss the update step. The prior step yielded a function $\hat{D}_{k|k-1}$ represented by $(\omega_{k|k-1}^{i},x_{k}^{i})_{i=1}^{L_{k}}$.~The update operator $\Psi_{k}(\hat{D}_{k|k-1}, Z_k)(x_{k})$ then maps this function into one with particle representation $(w_{k}^{i},x_{k}^{i})_{i=1}^{L_{k}}$:
\begin{equation*}
\begin{aligned}
\hat{D}_{k|k}(x)=\Psi_{k}(\hat{D}_{k|k-1}, Z_k)(x_{k})=\sum_{i=1}^{L_{k}}\omega_{k}^{i}\delta_{x_{k}^{i}}(x)\\
\end{aligned}
\end{equation*}
by modifying the weights of the particles as follows:
\begin{equation}\label{eq:SMC_PHD_Update}
\begin{aligned}
&\omega_{k}^{i}=\left[1-\pi(x_{k}^{i},q_{k})+\sum_{z\thinspace\in\thinspace Z_{k}}\frac{\psi_{kz}(x_{k}^{i})}{C_{k}(z)}\right]\omega_{k|k-1}^{i}
\end{aligned}
\end{equation}
where $C_{k}(z)=\sum_{j=1}^{L_{k}}\psi_{kz}(x_{k}^{j})\omega_{k|k-1}^{j}$.~Equation~\eqref{eq:SMC_PHD_Update} is a particle-based representation of~\eqref{eq:PHD_update}, and $\psi_{kz}$ is defined as discrete time of $\left<\psi_{kz},D_{k|k-1}\right>=\int_{E}\psi_{kz}(x_{k})D_{k|k-1}(x_{k})dx_{k}$.
\subsection*{Adapting Particle Numbers and Resampling Particles}\label{subsec:rsampling}
At any time step $k\geq1$ let $\hat{D}_{k|k}=\left\{\omega_{k}^{i},x_{k}^{i}\right\}_{i=1}^{L_{k}}$ denote a particle approximation of $D_{k|k}$, where $L_k$ is the particle count at $k$. The algorithm is designed such that the concentration of particles in a given region of the target space, say $S$, represents the approximated number of targets in $S$ depending on the sensor measurements. At times there may be too few or too many particles for a set of targets. It would be more efficient to adapt the allocation, say $\l$ particles per target at each time step $k$ where $\l$ is a tunable parameter. Since the expected number of targets $N_{k|k}$ is
\begin{equation}\label{eq:expected_card}
\hat{N}_{k|k}=\sum_{j=1}^{L_k}\omega_{k}^{j}
\end{equation}
it is natural to have the new number of particles $L^+_{k} = \l \hat{N}_{k|k}$.
Note that in this section, we use notations with superscript `+' for the particle count and weights after resampling; to keep notation manageable, in all other sections we simply use $L_k$, $\omega_{k}^{i}$ while leaving the resampling step implicit.

% all the L_ks and the omegas from the prior to resampling get renamed to tilde L k

To avoid the common problem of the weight variance increase, we use importance resampling.~The new particles are drawn randomly from the old set of particles with probabilities $a_i = \frac{\omega_{k}^{i}}{\sum_{j=1}^{L_k} \omega_{k}^{j}}$ and the weights are set as $\omega_{k}^{+i}=\frac{\hat{N}_{k|k}}{L^+_k}$ and thus sum up to $\hat{N}_{k|k}= \sum_{i=1}^{L_k}\omega_{k}^{i}$.
\end{appendices}

\newpage
\bibliography{sn-bibliography}% common bib file
\end{document}